\documentclass[letterpaper]{article} % DO NOT CHANGE THIS
\usepackage[]{aaai2026}
\usepackage{times}  % DO NOT CHANGE THIS
\usepackage{helvet}  % DO NOT CHANGE THIS
\usepackage{courier}  % DO NOT CHANGE THIS
\usepackage[hyphens]{url}  % DO NOT CHANGE THIS
\usepackage{graphicx} % DO NOT CHANGE THIS
\usepackage{natbib}  % DO NOT CHANGE THIS AND DO NOT ADD ANY OPTIONS TO IT
\usepackage{caption} % DO NOT CHANGE THIS AND DO NOT ADD ANY OPTIONS TO IT
\usepackage{booktabs} 
\usepackage{algorithm}
\usepackage{algorithmic}
\usepackage{amsmath}
\usepackage{amssymb}
\usepackage{amsfonts}
\DeclareMathOperator*{\argmin}{arg\,min}

\usepackage{newfloat}
\usepackage{listings}
\DeclareCaptionStyle{ruled}{labelfont=normalfont,labelsep=colon,strut=off} % DO NOT CHANGE THIS
\floatstyle{ruled}
\newfloat{listing}{tb}{lst}{}
\floatname{listing}{Listing}
\nocopyright
\title{Learning from B Cell Evolution: Adaptive Multi-Expert Diffusion for Antibody Design via Online Optimization}
\author{
   Hanqi Feng\textsuperscript{\rm 1}\equalcontrib,
   Peng Qiu\textsuperscript{\rm 1}\equalcontrib,
   Mengchun Zhang\textsuperscript{\rm 2}\equalcontrib,
   Yiran Tao\textsuperscript{\rm 1},
   You Fan\textsuperscript{\rm 3},
   Jingtao Xu\textsuperscript{\rm 4},
   Barnabas Poczos\textsuperscript{\rm 1}\thanks{Corresponding author}
}
\affiliations{
   \textsuperscript{\rm 1}School     of Computer Science, Carnegie Mellon University\\
   \textsuperscript{\rm 2}Department of Biostatistics, University of Pittsburgh\\
   \textsuperscript{\rm 3}Department of Mathematics, King's College London\\
   \textsuperscript{\rm 4}Department of Statistics, London School of Economics and Political Science\\
   \{hanqif, pengq, bpoczos\}@cs.cmu.edu, andy0814xu@gmail.com
}
\usepackage{todonotes}
\usepackage{bibentry}

\usepackage{etoolbox}
\makeatletter
\patchcmd{\maketitle}{\twocolumn[\@maketitle]}{\twocolumn[\vspace*{-10mm}\@maketitle]}{}{}
\makeatother

\begin{document}
\maketitle

\begin{abstract}
Recent advances in diffusion models have shown remarkable potential for antibody design, yet existing approaches apply uniform generation strategies that cannot adapt to each antigen's unique requirements. Inspired by B cell affinity maturation---where antibodies evolve through multi-objective optimization balancing affinity, stability, and self-avoidance---we propose the first biologically-motivated framework that leverages physics-based domain knowledge within an online meta-learning system. Our method employs multiple specialized experts (van der Waals, molecular recognition, energy balance, and interface geometry) whose parameters evolve during generation based on iterative feedback, mimicking natural antibody refinement cycles. Instead of fixed protocols, this adaptive guidance discovers personalized optimization strategies for each target. Our experiments demonstrate that this approach: 
(1) discovers optimal SE(3)-equivariant guidance strategies for different antigen classes without pre-training, preserving molecular symmetries throughout optimization; 
(2) significantly enhances hotspot coverage and interface quality through target-specific adaptation, achieving balanced multi-objective optimization characteristic of therapeutic antibodies; 
(3) establishes a paradigm for iterative refinement where each antibody-antigen system learns its unique optimization profile through online evaluation; 
(4) generalizes effectively across diverse design challenges, from small epitopes to large protein interfaces, enabling precision-focused campaigns for individual targets.
\end{abstract}


\section{Introduction}
Computational antibody design remains a fundamental challenge in therapeutic development, requiring simultaneous optimization of hotspot coverage, structural stability, and binding interface quality \cite{fischman2018computational, norman2020computational}. While recent diffusion-based methods like RFdiffusion \cite{watson2023novo} and RFAntibody \cite{luo2022} show promise for generating novel protein structures, they lack mechanisms to incorporate real-time constraints during generation, often producing designs that fail to meet multiple competing objectives \cite{eguchi2022igvae}.

Current approaches face three fundamental limitations: (1) they generate structures through diffusion without target-specific guidance \cite{watson2023novo, trippe2023diffusion}, (2) they cannot balance multiple objectives during generation, requiring inefficient post-hoc filtering \cite{jin2021iterative, shuai2021generative}, and (3) they either ignore physical constraints or require extensive labeled data to train property predictors \cite{dauparas2022robust, hsu2022learning}. These limitations significantly reduce their effectiveness for therapeutic antibody design, where each target presents unique challenges \cite{akbar2022progress, mason2021optimization}.

We present an adaptive guidance framework that addresses these limitations by introducing physics-aware constraints directly into the SE(3)-equivariant diffusion process \cite{hoogeboom2022equivariant, corso2023diffdock}. Our contributions include:
\begin{itemize}
    \item \textbf{Multi-expert guidance system}: We develop specialized guidance modules for van der Waals interactions \cite{jumper2021highly}, molecular recognition \cite{gainza2020deciphering}, energy balance \cite{ingraham2019generative}, and interface geometry \cite{schneider2022dlab}, each providing targeted gradients during diffusion while maintaining equivariance \cite{satorras2021en}.
    
    \item \textbf{Novel expert routing}: A dynamic routing system activates appropriate experts based on real-time structural metrics and diffusion timestep, ensuring optimal constraint application throughout generation \cite{xue2015computational, olympiou2022antibody}.
    
    \item \textbf{Online parameter adaptation}: Using Bayesian optimization with Gaussian processes, our framework learns optimal guidance strategies for each antigen-antibody pair through iterative batch evaluation, automatically discovering target-specific temporal profiles without requiring pre-training \cite{raybould2019five, shanehsazzadeh2023}.
\end{itemize}

Experiments across diverse targets demonstrate substantial improvements: 7\% reduction in CDR-H3 RMSD, 9\% increase in hotspot coverage, 12\% better interface pAE, and 5\% higher shape complementarity, with enhanced metrics across all evaluation criteria. This balanced optimization addresses the critical “weakest link” problem~\cite{raybould2019five} in antibody design. By combining physics-based guidance with online learning, our work opens new directions for adaptive approaches in biomolecular design.
\section{Related Work}
\subsection{Computational Antibody Design}
The field has evolved from physics-based approaches using Rosetta \cite{kaufmann2010practically} and MODELLER \cite{webb2016comparative} to deep learning methods inspired by AlphaFold2 \cite{jumper2021highly}. DeepAb \cite{ruffolo2021deepab} pioneered deep learning for antibody structure prediction, while ABlooper \cite{abanades2022ablooper} focused on CDR-H3 modeling.

Recent generative approaches shifted to de novo design. DiffAb \cite{luo2022} introduced diffusion models for joint sequence-structure generation but operates in internal coordinates, limiting inter-chain modeling. RFAntibody \cite{adolf2024denovo} addressed this using SE(3)-equivariant backbone generation in Cartesian space. However, current methods lack both physicochemical constraint enforcement and target-specific adaptation during generation. This results in high-throughput campaigns producing substantial fractions of structurally unreasonable or experimentally nonviable designs \cite{shen2024chemistry}. Additional related work is discussed in Appendix A due to space limitations.

\section{Preliminaries}

\subsection{SE(3)-Equivariant Diffusion Models}

Protein structure modeling requires respecting 3D spatial symmetries. RFdiffusion \cite{watson2023novo} combines SE(3)-equivariant networks with diffusion models, performing diffusion directly on backbone coordinates while maintaining rotational and translational equivariance.

\subsubsection{Frame Representations and Forward Diffusion Process}

Each residue $i$ is represented by a rigid body frame $\mathbf{T}_i = (\mathbf{R}_i, \mathbf{t}_i) \in \text{SE}(3)$, where $\mathbf{R}_i \in \text{SO}(3)$ is a rotation matrix and $\mathbf{t}_i \in \mathbb{R}^3$ is a translation vector. For $N$ residues, the complete structure is $\mathbf{T} = \{\mathbf{T}_1, \ldots, \mathbf{T}_N\}$.

The forward diffusion process adds noise over time steps $t \in \{0, 1, \ldots, T\}$. Rotations follow the IGSO3 distribution, $q(\mathbf{R}_t|\mathbf{R}_0) = \text{IGSO3}(\mathbf{R}_t; \mathbf{R}_0, \sigma_t)$ with density:
\begin{align}
f(\mathbf{R}_t|\mathbf{R}_0, \sigma_t) &= \frac{1}{Z(\sigma_t)} \exp\left(\frac{\text{tr}(\mathbf{R}_0^T\mathbf{R}_t)}{\sigma_t^2}\right)
\end{align}
where $\sigma_t$ is the noise level at timestep $t$, controlling the variance of the distribution\cite{yim2023se}. The derivation of the score function and its properties on the SO(3) manifold are detailed in Appendix B.
Translations follow Gaussian diffusion with center-of-mass constraint:
\begin{equation}
q(\mathbf{t}_t|\mathbf{t}_0) = \mathcal{N}(\mathbf{t}_t; \mathbf{t}_0, \sigma_t^2\mathbf{I}_3), \quad \text{s.t.} \quad \sum_{i=1}^N \mathbf{t}_{t,i} = \sum_{i=1}^N \mathbf{t}_{0,i}
\end{equation}

\textbf{Reverse Diffusion and Network Architecture} The reverse process learns to denoise the forward diffusion process by estimating clean structure $\mathbf{T}_0$ from noisy $\mathbf{T}_t$ \cite{ho2020denoising}:
\begin{align}
p_\theta(\mathbf{T}_{t-1}|\mathbf{T}_t) &= p(\mathbf{T}_{t-1}|\mathbf{T}_t, \hat{\mathbf{T}}_0), \\
\hat{\mathbf{T}}_0 &= \frac{1}{\sqrt{\bar{\alpha}_t}}\left(\mathbf{T}_t - \sqrt{1-\bar{\alpha}_t}\epsilon_\theta(\mathbf{T}_t, t)\right)
\end{align}
where $\bar{\alpha}_t = \prod_{s=1}^{t} \alpha_s$ is the cumulative product of noise schedule parameters, and $\epsilon_\theta(\mathbf{T}_t, t)$ is the neural network with parameters $\theta$ that predicts the noise added at timestep $t$.

\subsection{RFAntibody Pipeline}
RFAntibody~\cite{adolf2024rfantibody} employs a three-stage pipeline for antibody design: (1) RFdiffusion generates diverse antibody backbones conditioned on the target antigen, (2) ProteinMPNN designs sequences compatible with these structures, and (3) RoseTTAFold2 validates the designs by predicting their folded structures. This decoupled approach allows each model to focus on its specialized task—structure generation, sequence design, and validation respectively. The pipeline filters designs based on structural quality metrics, retaining only those most likely to succeed experimentally.

\section{Problem Formulation and Motivation}

While state-of-the-art generative models excels at generating diverse protein structures, it faces a critical challenge in antibody design: maintaining balanced performance across multiple essential metrics. In antibody engineering, success requires simultaneous optimization of structural accuracy (CDR-H3 RMSD), prediction confidence (pAE, ipAE), interface quality (shape complementarity, buried surface area), and biophysical properties (VDW interactions, stability) \cite{raybould2019five, norman2020computational, choi2018predicting}.

The fundamental issue is that RFdiffusion's purely data-driven approach lacks mechanisms to enforce multi-objective balance. This leads to the "weakest link" problem—a single failing metric renders the entire design experimentally nonviable~\cite{raybould2019five,choi2018predicting}, regardless of excellence in other areas. In high-throughput campaigns, even if 90\% of metrics are excellent, failure in the remaining 10\% eliminates the design from experimental consideration.

\begin{figure}[h]
    \centering
    \includegraphics[width=\columnwidth]{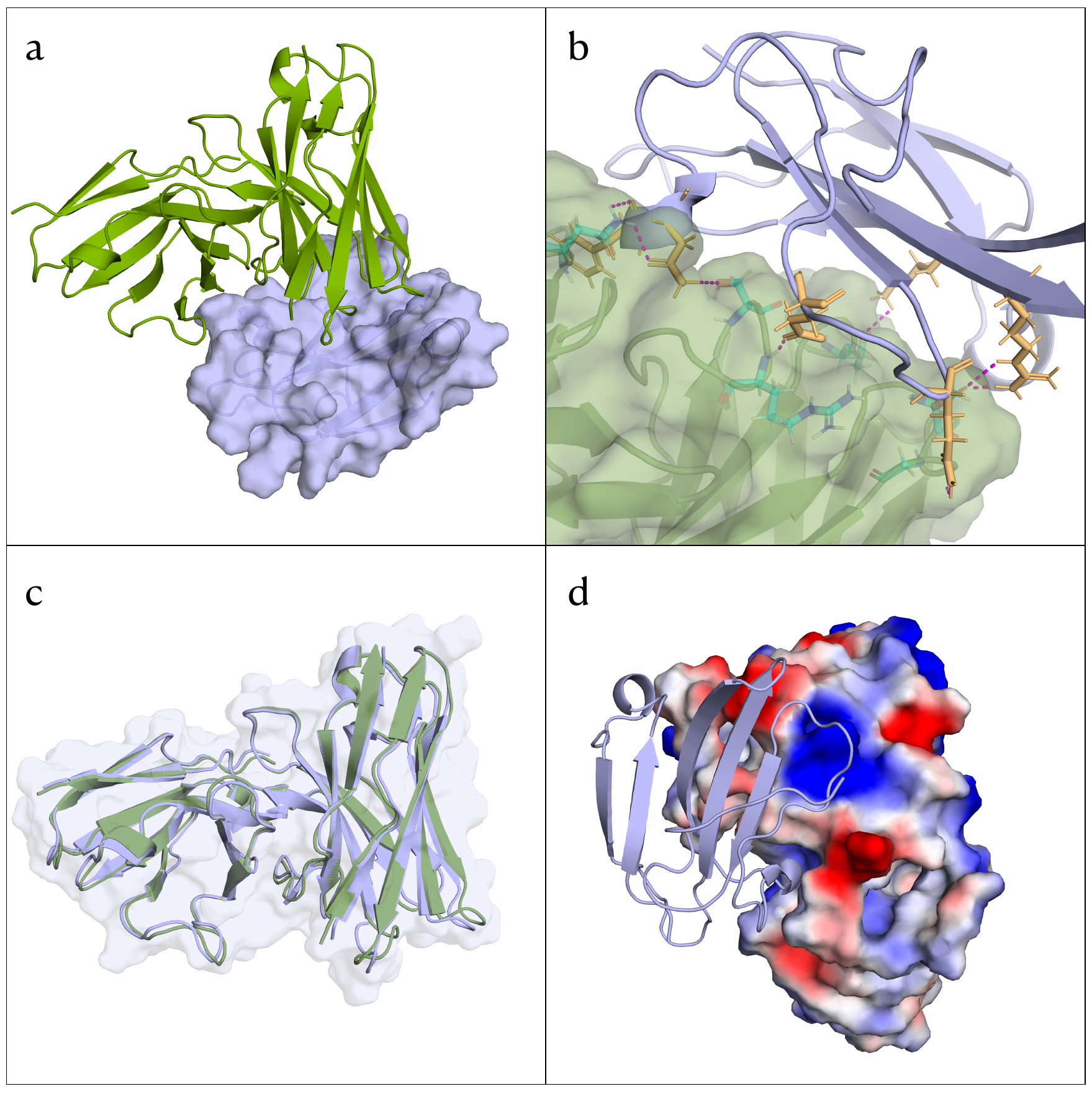}
    \caption{High-quality antibody design showcasing the effectiveness of our adaptive guidance. 
    (a) Overview of the antibody-antigen complex. 
    (b) Detailed view of the binding interface demonstrating enhanced contact density achieved through guided optimization. 
    (c) Structural alignment showing exceptional CDR-H3 accuracy (RMSD = 0.63 \AA). 
    (d) Electrostatic surface representation illustrating improved charge complementarity resulting from physics-aware guidance.}
    \label{fig:design_showcase}
\end{figure}

Our work addresses this challenge by developing an adaptive guidance system that steers the diffusion process toward regions where all objectives are simultaneously satisfied. Unlike post-hoc filtering approaches that discard failed designs, our method proactively guides generation to maintain metric balance throughout the process, significantly improving the yield of experimentally viable candidates per computational cycle (Figure~\ref{fig:design_showcase}).

\section{Method}

\subsection{Theoretical Foundation}

\subsubsection{Biophysical Principles of Antibody-Antigen Recognition}
Our multi-expert system is grounded in the fundamental physics of molecular recognition. Antibody-antigen binding emerges from a delicate balance of forces:

\textbf{Energetic Landscape:} The binding free energy ($\Delta G_{\text{bind}}$) can be decomposed as ~\cite{gilson2007calculation}: 
\begin{equation}
\Delta G_{\text{bind}} = \Delta H_{\text{vdW}} + \Delta H_{\text{elec}} + \Delta H_{\text{hbond}} - T\Delta S_{\text{conf}} + \Delta G_{\text{solv}}
\end{equation}
where each term corresponds to van der Waals, electrostatic, hydrogen bonding, conformational entropy, and solvation contributions respectively. Our expert modules directly optimize these physical components.

\textbf{Shape Complementarity Principle:} Following the lock-and-key model extended by induced fit theory \cite{fischer1894einfluss, koshland1958application}, optimal binding requires:
\begin{equation}
SC = \frac{1}{2}\left(\frac{S_{\text{buried,Ab}}}{S_{\text{total,Ab}}} + \frac{S_{\text{buried,Ag}}}{S_{\text{total,Ag}}}\right) \cdot \exp\left(-\frac{d_{\text{gap}}}{\lambda}\right)
\end{equation}
where Ab and Ag denote antibody and antigen respectively, $S_{\text{buried}}$ represents buried surface area upon complex formation, $S_{\text{total}}$ is the total surface area of each protein, and $d_{\text{gap}}$ captures interface gaps penalized by decay length $\lambda$.

\subsubsection{Information-Theoretic Justification for Adaptive Guidance}
Our adaptive expert routing is motivated by information theory. At timestep $t$, the mutual information between the noisy structure $\mathbf{x}_t$ and the target structure $\mathbf{x}_0$ is~\cite{austin2021structured}:
\begin{equation}
I(\mathbf{x}_t; \mathbf{x}_0) = H(\mathbf{x}_0) - H(\mathbf{x}_0|\mathbf{x}_t) \propto \frac{\alpha_t}{\sigma_t^2}
\end{equation}
where $\alpha_t$ is the signal retention coefficient at timestep $t$. The derivation of this relationship is provided in Appendix C.

This suggests that different structural features become identifiable at different noise levels, justifying our adaptive expert routing based on the signal-to-noise ratio at each timestep. The connection between SNR and feature visibility is detailed in Appendix D.

\subsection{Overview}
\begin{figure*}
   \centering
   \includegraphics[width=1\linewidth]{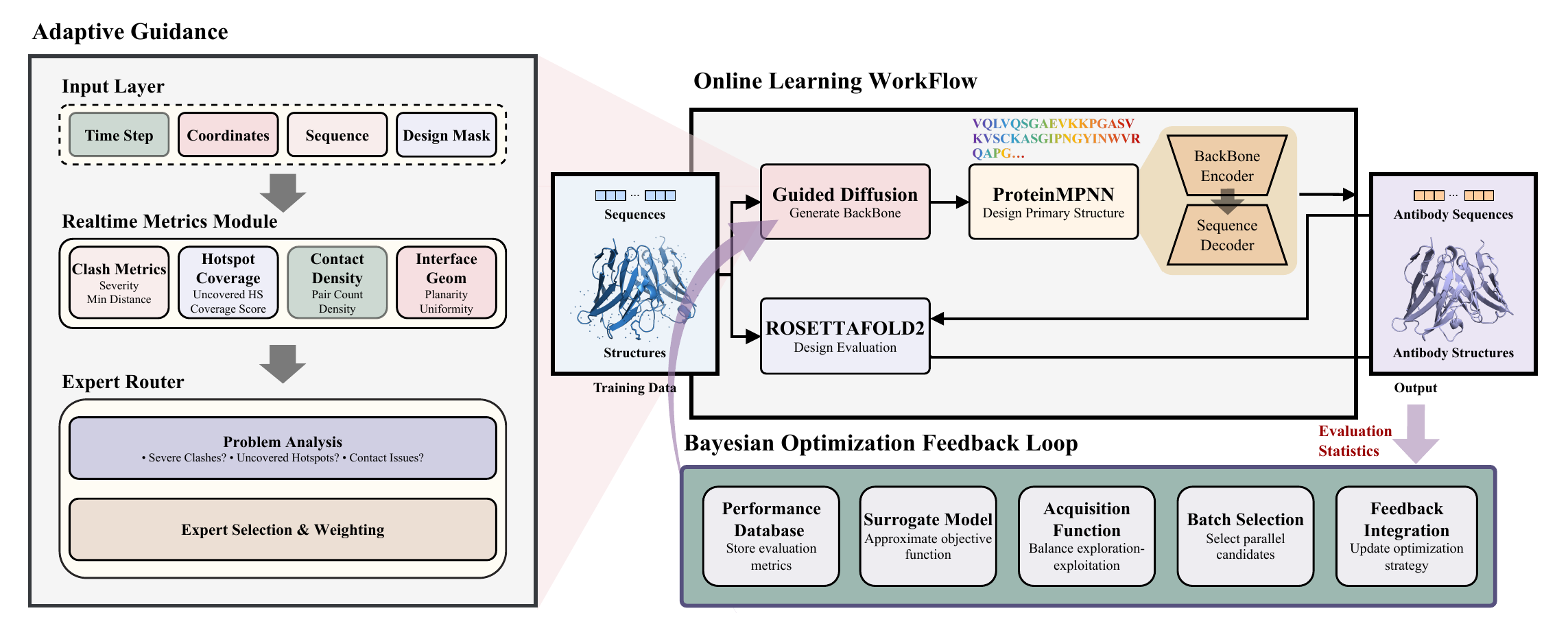}
   \caption{Adaptive multi-expert diffusion framework with online learning for antibody design. Due to space constraints, detailed descriptions of each module and their input/output dimensions are provided in Appendix F.}
   \label{fig:pipeline}
\end{figure*}

We present an adaptive physics-guided diffusion framework for antibody design that integrates multi-expert guidance with online learning. Our approach enhances the standard SE(3) equivariant diffusion process \cite{hoogeboom2022equivariant, watson2023novo} by incorporating domain-specific physical constraints through a dynamically weighted ensemble of expert modules. Figure~\ref{fig:pipeline} presents our adaptive framework. The system combines physics-based expert guidance (left) with the antibody design pipeline, where guided diffusion generates structures, ProteinMPNN designs sequences, and RoseTTAFold2 validates results. A Bayesian optimization feedback loop (bottom) continuously learns optimal guidance parameters from evaluation metrics.

\subsection{Physics-Guided SE(3) Diffusion}

Our framework modifies the standard reverse diffusion process by incorporating physics-based guidance gradients. The reverse step is formulated as:
\begin{equation}
    \mathbf{T}_{t-1} = \boldsymbol{\mu}_\theta(\mathbf{T}_t, t) + \boldsymbol{\Sigma}_\theta^{1/2}(\mathbf{T}_t, t) \cdot \mathbf{z} - \lambda(t) \cdot \mathbf{g}(\mathbf{T}_t, t)
\end{equation}
where $\boldsymbol{\mu}_\theta$ and $\boldsymbol{\Sigma}_\theta$ are the predicted mean and covariance from the denoising network, $\mathbf{z} \sim \mathcal{N}(0, \mathbf{I})$, and $\mathbf{g}(\mathbf{T}_t, t)$ represents the combined guidance gradient:
\begin{equation}
    \mathbf{g}(\mathbf{T}_t, t) = \sum_{i=1}^{4} w_i(t) \cdot \lambda_i(t) \cdot \nabla \mathcal{L}_i(\mathbf{T}_t)
\end{equation}
where $w_i(t)$ are adaptive weights for each expert and $\nabla\mathcal{L}_i$ are gradients of expert-specific physics-based loss functions.

The guidance strength for each expert $i$ at timestep $t$ is determined as:
\begin{equation}
    \lambda_i(t) = \lambda_{\text{base},i} \cdot f_{\text{temporal}}(t, \alpha, \beta)
\end{equation}
where $\lambda_{\text{base},i}$ is the base strength for expert $i$, and $f_{\text{temporal}}(t, \alpha, \beta)$ controls when each expert is most active during denoising. The shape parameters $\alpha$ and $\beta$ are learned via 
Bayesian optimization to discover optimal temporal profiles—early-peaking 
($\alpha < \beta$) for global structure experts, late-peaking ($\alpha > \beta$) 
for atomic refinement:

\begin{equation}
   f_{\text{temporal}}(t, \alpha, \beta) = \frac{B(t_{\text{norm}}; \alpha, \beta)}{B(t_{\text{mode}}; \alpha, \beta)} \cdot \lambda_{\text{peak}}
\end{equation}

where $t_{\text{norm}} = (t-1)/(T-1)$ normalizes the timestep to $[0,1]$, and $B(\cdot; \alpha, \beta)$ is the Beta density function. The shape parameters $\alpha$ and $\beta$ are learned via Bayesian optimization to discover optimal activation timing for each target. Detailed profiles are shown in Appendix E.

\subsection{Multi-Expert Guidance System}

We employ four specialized expert modules, each addressing critical aspects of antibody-antigen interactions:

\subsubsection{VDW Balance Expert}
This expert prevents steric clashes while maintaining favorable van der Waals interactions. The loss function penalizes atomic overlaps:

\begin{equation}
\mathcal{L}_{\text{vdw}} = \sum_{ \{i,j\} \subseteq \text{CDR} \cup \text{target}} \max(0, r_{\text{clash}} - d_{ij})^2
\end{equation}

where $r_{\text{clash}}$ is the clash threshold and $d_{ij} = \|\mathbf{x}_i - \mathbf{x}_j\|$ is the distance between backbone atoms. The gradient is:

\begin{equation}
    \nabla \mathcal{L}_{\text{vdw}} = 2 \sum_{d_{ij} < r_{\text{clash}}} \frac{r_{\text{clash}} - d_{ij}}{d_{ij}} \cdot (\mathbf{x}_i - \mathbf{x}_j)
\end{equation}

\subsubsection{Molecular Recognition Expert}
This expert ensures proper CDR coverage of epitope hotspots. For each hotspot $h \in \mathcal{H}$, we encourage proximity to CDR residues(see Appendix G for hotspot definition and selection methodology):

\begin{equation}
    \mathcal{L}_{\text{hotspot}} = \sum_{h \in \mathcal{H}} \min_{c \in \text{CDR}} \|\mathbf{x}_h - \mathbf{x}_c\|^2
\end{equation}

The gradient attracts the nearest CDR residue $c^*$ toward uncovered hotspots:

\begin{equation}
    \nabla \mathcal{L}_{\text{hotspot}} = 2 \sum_{h \in \mathcal{H}_{\text{uncovered}}} (\mathbf{x}_{c^*} - \mathbf{x}_h), \quad c^* = \argmin_{c \in \text{CDR}} \|\mathbf{x}_h - \mathbf{x}_c\|
\end{equation}

where $\mathcal{H}_{\text{uncovered}}$ denotes inadequately covered hotspots.

\subsubsection{Energy Balance Expert}
This expert maintains optimal contact density at the binding interface. The loss function is defined as:

\begin{equation}
    \mathcal{L}_{\text{contact}} = 
    \begin{cases}
        \phi^{-}(n_c) & \text{if } n_c < \tau^{-} \\
        \phi^{+}(n_c) & \text{if } n_c > \tau^{+} \\
        0 & \text{if } \tau^{-} \leq n_c \leq \tau^{+}
    \end{cases}
\end{equation}

where $n_c$ denotes the number of inter-molecular contacts within threshold $d_c$, $[\tau^{-}, \tau^{+}]$ defines the target range, and $\phi^{-}(\cdot)$, $\phi^{+}(\cdot)$ are penalty functions for sparse and dense packing, respectively. The gradient modulates CDR positioning to optimize interface contacts.

\subsubsection{Interface Quality Expert}
This expert optimizes the geometric properties of the binding interface through multiple physically motivated objectives:

\begin{equation}
    \mathcal{L}_{\text{geom}} = w_u \cdot \mathcal{L}_{\text{uniformity}} + w_c \cdot \mathcal{L}_{\text{cavity}}
\end{equation}

where $w_u$ and $w_c$ are weighting coefficients that balance the contribution of each geometric criterion.

\textbf{Distance Uniformity:} The uniformity of residue spacing is evaluated to ensure well-distributed interface contacts:

\begin{equation}
    \mathcal{L}_{\text{uniformity}} = \frac{\sigma(d_{ij})}{\mu(d_{ij})}, \quad \forall (i,j) \in \mathcal{I}, \; d_{ij} < d_{\text{cutoff}}
\end{equation}

where $d_{ij}$ represents pairwise distances between interface residues $\mathcal{I}$, and $\sigma(\cdot)$, $\mu(\cdot)$ denote standard deviation and mean, respectively. This metric prevents local clustering or sparse regions at the interface.

\textbf{Cavity Detection:} The cavity penalty identifies isolated residues lacking sufficient neighbors:

\begin{equation}
    \mathcal{L}_{\text{cavity}} = \frac{1}{|\mathcal{I}|} \sum_{i \in \mathcal{I}} \mathbf{1}[\mathcal{N}_i < n_{\text{threshold}}]
\end{equation}

where $\mathcal{N}_i$ counts neighbors within radius $r_{\text{neighbor}}$, and $\mathbf{1}[\cdot]$ is the indicator function that equals 1 if the condition is true and 0 otherwise. This penalty prevents the formation of hydrophobic cavities and ensures a well-packed interface.

\subsection{Adaptive Expert Routing}

Our routing mechanism dynamically activates experts based on real-time structural analysis, ensuring that computational resources focus on the most critical problems while maintaining SE(3)-equivariance. The system computes problem severity scores:
\begin{equation}
   s_i = f_i(\text{metrics}_t) \in [0, 1]
\end{equation}
where $f_i$ evaluates current structural metrics to determine the severity of problems relevant to expert $i$ (detailed in the Appendix H).

The activation is problem-driven rather than time-dependent. For instance:
\begin{itemize}
   \item High clash density ($d_{\min} < r_{\text{clash}}$) triggers VDW expert activation
   \item Low hotspot coverage activates molecular recognition guidance  
   \item Suboptimal contact density engages energy balance optimization
   \item Poor interface geometry activates geometric refinement
\end{itemize}

Experts are weighted proportionally to problem severity:
\begin{equation}
   w_i(t) = \frac{s_i}{\sum_{j: s_j > \theta_{\min}} s_j} \cdot \mathbf{1}[s_i > \theta_{\min}]
\end{equation}
This adaptive weighting ensures that only experts addressing detected problems are activated, preventing unnecessary or conflicting guidance.

Critically, SE(3)-equivariance is preserved because all expert gradients are computed from invariant geometric features. Each expert's gradient satisfies:
\begin{equation}
   \nabla\mathcal{L}_i(\mathbf{R}\mathbf{T} + \mathbf{t}) = \mathbf{R}\nabla\mathcal{L}_i(\mathbf{T})
\end{equation}
for any rotation $\mathbf{R} \in \text{SO}(3)$ and translation $\mathbf{t} \in \mathbb{R}^3$. This is achieved by basing all computations on pairwise distances, relative orientations, and other invariant descriptors rather than absolute coordinates. The weighted combination of equivariant gradients maintains equivariance: $\mathbf{g}(\mathbf{T}_t, t) = \sum_i w_i(t) \cdot \nabla\mathcal{L}_i(\mathbf{T}_t)$ remains equivariant since weights $w_i(t)$ are scalar functions of invariant metrics.

\subsection{Online Parameter Learning}

We employ Bayesian optimization with Gaussian processes (GP) to adaptively learn the optimal Beta distribution parameters for each antibody-antigen system post-generation. Our approach models a two-dimensional parameter space $\boldsymbol{\theta} = (\alpha, \beta) \in \Theta$, where $\alpha$ and $\beta$ are the shape parameters controlling the temporal modulation profile.
The optimization process maintains a probabilistic surrogate model of the objective function using a Gaussian process:
\begin{equation}
   f(\boldsymbol{\theta}) \sim \mathcal{GP}(\mu(\boldsymbol{\theta}), k(\boldsymbol{\theta}, \boldsymbol{\theta}'))
\end{equation}
where $\mu(\boldsymbol{\theta})$ represents the predicted mean performance and $k(\boldsymbol{\theta}, \boldsymbol{\theta}')$ is the covariance function. We employ a Matérn kernel with automatic noise estimation:
\begin{equation}
   k(\boldsymbol{\theta}, \boldsymbol{\theta}') = k_{\text{Matérn}}(\boldsymbol{\theta}, \boldsymbol{\theta}') + \sigma_n^2(\boldsymbol{\theta})\delta_{ij}
\end{equation}

After each design evaluation, the GP posterior is updated using the observed loss:
\begin{equation}
   L(\boldsymbol{\theta}) = \sum_{m=1}^{M} \omega_m \cdot \frac{\text{metric}_m}{\nu_m}
\end{equation}
where the metrics include CDR-H3 backbone RMSD, predicted aligned error (pAE), and interaction pAE (ipAE), each normalized by appropriate constants $\nu_m$.

The next parameter configuration is selected by maximizing the Expected Improvement (EI) acquisition function:
\begin{equation}
   \boldsymbol{\theta}_{n+1} = \arg\max_{\boldsymbol{\theta} \in \Theta} \text{EI}(\boldsymbol{\theta})
\end{equation}

To handle stochastic evaluations, we aggregate results from similar parameters to reduce noise and periodically re-evaluate promising configurations. This approach automatically discovers optimal Beta parameters for each target without manual tuning. Due to space constraints, convergence analysis and computational requirements are detailed in Appendix I.

\section{Experiments}

\subsection{Experimental Setup}

\subsubsection{Dataset and Targets}
We evaluated our framework on six antibody-antigen pairs: IL17A (PDB: 6PPG), ACVR2B (PDB: 5NGV), FXI (PDB: 6HHC), TSLP (PDB: 5J13), IL36R (PDB: 6U6U), and TNFRSF9 (PDB: 6A3W). Prior work including RFAntibody~\cite{bennett2024} typically evaluated four targets; we expanded this to six therapeutically relevant targets spanning inflammation, thrombosis, and cancer applications (17-40 kDa). This extended benchmark from IgDesign~\cite{shanehsazzadeh2023} includes 1,243 SPR-validated designs across diverse binding modes. We annotated 5 hotspot residues per target and generated 1,000 designs each, following the field's established protocol for rigorous benchmarking of antibody design methods.

\subsubsection{Baseline Methods}
Most recent methods are unavailable due to lack of publicly accessible code. We selected: \textbf{DiffAb}~\citep{luo2022}: torsion-space CDR generation; \textbf{RFAntibody}~\citep{bennett2024}: the current state-of-the-art method without physics guidance, widely recognized for its exceptional performance across diverse antibody design tasks.  For fair comparison, all methods used identical inputs (target structure, hotspot annotations, and Trastuzumab framework).
\subsubsection{Implementation Details}
All experiments used 4 NVIDIA RTX A6000 GPUs with the following settings in Table~\ref{tab:hyperparams}.

\begin{table}[h]
\centering
\small
\setlength{\tabcolsep}{2pt}  % 减小列间距
\begin{tabular*}{\columnwidth}{@{\extracolsep{\fill}}ll|ll@{}}
\hline
\textbf{Parameter} & \textbf{Value} & \textbf{Parameter} & \textbf{Value} \\
\hline
\multicolumn{2}{l|}{\textit{Diffusion \& Guidance}} & \multicolumn{2}{l}{\textit{Beta Distribution}} \\
Timesteps $T$ & 50 & $\alpha$ range & [0.5, 10.0] \\
Expert modules & 4 & $\beta$ range & [0.5, 10.0] \\
VDW threshold & 2.8 Å & Peak factor $\lambda_{\text{peak}}$ & 5.0 \\
Gradient clip & 2.0 & Initial $(\alpha, \beta)$ & (2.0, 2.0) \\
Distance cutoff & 8.0 Å & & \\
\hline
\multicolumn{2}{l|}{\textit{Expert Activation}} & \multicolumn{2}{l}{\textit{Online Learning}} \\
Min threshold $\theta_{\min}$ & 0.1 & Kernel & Matérn-5/2 \\
VDW strength & 0.5 & Length scale & 2.0 \\
Recognition strength & 2.0 & GP noise & 0.3 \\
Activation & Adaptive & Acquisition $\xi$ & 0.01 \\
\hline
\multicolumn{2}{l|}{\textit{Evaluation Metrics}} & \multicolumn{2}{l}{\textit{Implementation}} \\
RMSD norm $\nu_1$ & 1.5 Å & Framework & PyTorch \\
pAE norm $\nu_2$ & 7.0 & GPU memory & 48GB \\
ipAE norm $\nu_3$ & 10.0 & CDR design & All 6 \\
\hline
\end{tabular*}
\caption{Implementation details of our adaptive physics-guided framework}
\label{tab:hyperparams}
\end{table}

\subsection{Evaluation Metrics}

We employ a comprehensive set of metrics to evaluate antibody design quality across structural accuracy, binding interface characteristics, and biophysical properties.

\subsubsection{CDR-H3 RMSD:} We measure the backbone root-mean-square deviation of the critical CDR-H3 loop after structural alignment of framework regions. CDR-H3 is the most diverse and challenging loop to model, directly correlating with binding specificity. 

\subsubsection{Predicted Alignment Error (pAE):} Extracted from AlphaFold2-Multimer confidence predictions, pAE estimates the expected position error between residue pairs. We report both mean pAE across the entire complex and mean interaction pAE (ipAE) specifically for residue pairs across the antibody-antigen interface. Lower values indicate higher confidence in the predicted structure.

\subsubsection{Predicted Local Distance Difference Test (pLDDT):} Per-residue confidence metric. We average pLDDT across CDR residues for local structural quality assessment.

\subsubsection{Shape Complementarity (SC):} Quantifies geometric fit between antibody-antigen surfaces, with higher values indicating better complementarity.

\subsubsection{Buried Surface Area (BSA):} Solvent-accessible surface area buried upon complex formation. Larger interfaces correlate with stronger, more specific interactions.

\subsubsection{Hotspot Coverage:} Percentage of critical epitope residues forming close contacts with CDR residues.

\subsubsection{CDR Participation:} Fraction of CDR residues involved in antigen binding. Higher participation indicates more distributed binding.

\subsubsection{Energetic and Biophysical Metrics}
~\\
\textbf{Van der Waals Energy:} We compute packing interactions and steric clashes using physics-based energy functions. Favorable scores indicate well-packed interfaces without significant clashes.

\subsubsection{Overall Success Criteria:}
Successful designs must satisfy: CDR-H3 RMSD  $<$ 3.0 Å, mean pAE  $<$ 10.0, and mean interaction pAE (ipAE)  $<$ 10.0. This multi-metric evaluation ensures structural accuracy and binding confidence.

This multi-metric evaluation ensures designs are not only structurally accurate but also likely to bind with high affinity and specificity. 

\begin{figure}[h]
   \centering
   \includegraphics[width=1\linewidth]{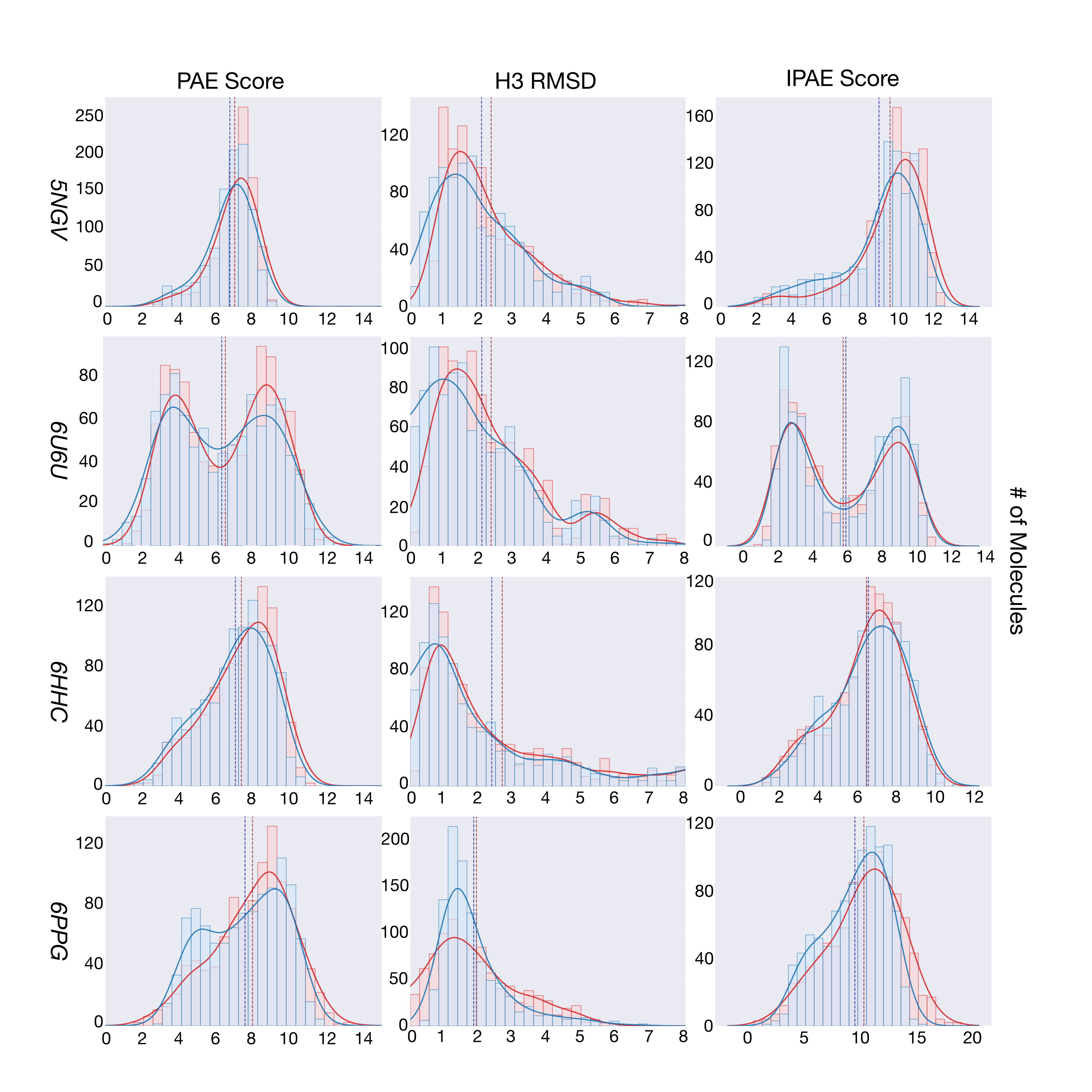}
   \caption{Histogram distributions of evaluation metrics (pAE, CDR-H3 RMSD, ipAE) for baseline (red) and adaptive guidance (blue) methods across four antibody targets.}
   \label{fig:metric-distributions}
\end{figure}

\subsection{Main Results}
\begin{figure}
    \centering
    \includegraphics[width=1\linewidth]{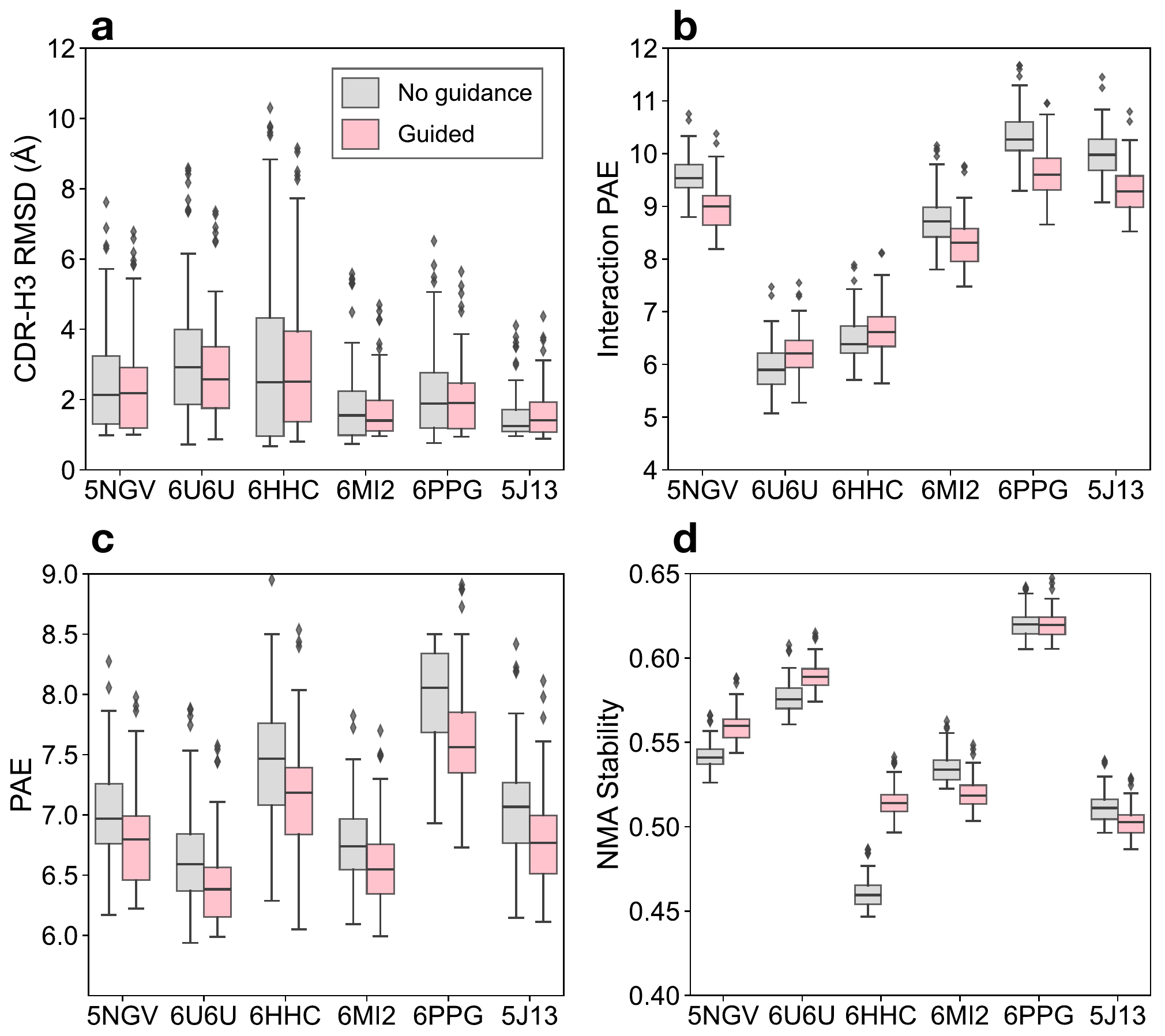}
    \caption{ Adaptive guidance improves antibody design quality across multiple metrics. Box plots compare baseline RFAntibody (gray) with our guided approach (pink) on six targets for (a) CDR-H3 RMSD, (b) interaction pAE, (c) mean pAE, and (d) NMA stability.}
    \label{fig:box}
\end{figure}

Table~\ref{tab:overall_multi} presents comprehensive evaluation across all targets. Our method achieved improvements in most metrics while maintaining a more balanced performance profile compared to both baselines, effectively addressing the critical ``weakest link'' problem in antibody design where poor performance in any single metric can compromise experimental viability. As shown in Figure~\ref{fig:metric-distributions} and Figure~\ref{fig:box}, our adaptive guidance consistently improves performance across all key metrics (CDR-H3 RMSD, pAE, ipAE) for multiple targets, with notably reduced variance in the distributions. 

\begin{table}[!h]
\centering
\small
\setlength{\tabcolsep}{0pt}
\begin{tabular*}{\columnwidth}{@{\extracolsep{\fill}}lcccc@{}}
\toprule
Method & Success(\%) & H3-RMSD(\AA) & pAE & iPAE \\
\midrule
DiffAb     & 11.1$\pm$17.0 & 9.1$\pm$10.1  & 9.4$\pm$1.2 & 10.1$\pm$2.2 \\
RFAntibody & 36.7$\pm$14.0 & 2.1$\pm$1.8   & 7.1$\pm$0.5 & 8.5$\pm$1.7  \\
Ours       & \textbf{41.1$\pm$10.5} & \textbf{1.9$\pm$1.4} & \textbf{6.9$\pm$0.4} & \textbf{8.1$\pm$1.3} \\
\midrule
Method & SC & BSA & VDW(kcal/mol) & pLDDT \\
\midrule
DiffAb     & 0.55$\pm$0.14 & \textbf{5277$\pm$2341} & 15.2$\pm$9.7  & 0.886$\pm$0.01 \\
RFAntibody & 0.55$\pm$0.03 & 3583$\pm$888  & $-$1.2$\pm$1.3 & \textbf{0.896$\pm$0.01} \\
Ours       & \textbf{0.58$\pm$0.02} & 3870$\pm$780 & \textbf{$-$1.3$\pm$1.0} & 0.893$\pm$0.01 \\
\midrule
Method & Hotspot(\%) & CDR Int.(\%) & Seq. Div. & Elec. Energy \\
\midrule
DiffAb     & 16.2$\pm$5.5  & 0.0$\pm$0.0   & 0.46$\pm$0.08 & \textbf{$-$6.3$\pm$11.1} \\
RFAntibody & 48.5$\pm$18.1 & 54.7$\pm$13.6 & \textbf{0.53$\pm$0.00} & $-$4.0$\pm$7.1 \\
Ours       & \textbf{57.5$\pm$12.6} & \textbf{61.3$\pm$11.2} & 0.53$\pm$0.00 & $-$5.6$\pm$6.1 \\
\bottomrule
\end{tabular*}
\caption{Overall performance comparison across six targets highlighting balanced optimization}
\label{tab:overall_multi}
\end{table}
\subsubsection{Online Learning Impact}
To evaluate the contribution of our online learning framework, we conducted ablation studies comparing three configurations: (1) the full system with both multi-expert guidance and online learning, (2) multi-expert guidance with fixed parameters (no online learning), and (3) the baseline RFAntibody without any guidance.

\begin{table}[h]
\centering
\small
\setlength{\tabcolsep}{1.5pt}  % 进一步减小列间距
\begin{tabular*}{\columnwidth}{@{\extracolsep{\fill}}lcccc@{}}
\toprule
Config & Success & H3-RMSD & pAE & iPAE \\
\midrule
RFdiffusion & 36.7$\pm$14.0 & 2.11$\pm$1.81 & 7.13$\pm$0.46 & 8.49$\pm$1.71 \\
+ Expert & 38.9$\pm$12.5 & 2.01$\pm$1.58 & 6.95$\pm$0.42 & 8.28$\pm$1.65 \\
\textbf{+ Full} & \textbf{41.1$\pm$10.5} & \textbf{1.94$\pm$1.36} & \textbf{6.85$\pm$0.39} & \textbf{8.12$\pm$1.31} \\
\midrule
Config & SC & BSA & VDW & pLDDT \\
\midrule
RFdiffusion & 0.549$\pm$.026 & 3583$\pm$888 & $-$1.18$\pm$1.27 & 0.896$\pm$.011 \\
+ Expert & 0.562$\pm$.026 & 3712$\pm$885 & $-$1.25$\pm$1.15 & 0.885$\pm$.014 \\
\textbf{+ Full} & \textbf{0.576$\pm$.024} & \textbf{3870$\pm$780} & \textbf{$-$1.33$\pm$0.95} & \textbf{0.893$\pm$.008} \\
\midrule
Config & Hotspot & CDR & Seq.D & Elec.E \\
\midrule
RFdiffusion & 48.5$\pm$18.1 & 54.7$\pm$13.6 & 0.534$\pm$.004 & $-$3.96$\pm$7.1 \\
+ Expert & 55.6$\pm$15.2 & 60.8$\pm$13.5 & 0.532$\pm$.005 & $-$5.31$\pm$8.1 \\
\textbf{+ Full} & \textbf{57.5$\pm$12.6} & \textbf{61.3$\pm$11.2} & \textbf{0.532$\pm$.004} & \textbf{$-$5.64$\pm$6.1} \\
\bottomrule
\end{tabular*}
\caption{Ablation study demonstrating the importance of online learning}
\label{tab:ablation}
\end{table}
The results reveal the complementary benefits of multi-expert guidance and online learning:

\textbf{Multi-Expert Guidance Alone:} Adding physics-based guidance to RFAntibody improves the success rate from 36.7\% to 38.9\% (Table~\ref{tab:ablation}). While all metrics show improvement, the gains are modest and the standard deviations remain relatively high. This suggests that fixed guidance parameters, though helpful, cannot adapt to the diverse characteristics of different antibody-antigen systems.

\textbf{Online Learning Enhancement:} The addition of online parameter optimization yields substantial improvements across all metrics. Most notably, the standard deviations decrease significantly, indicating more consistent and reliable design quality. This variance reduction is crucial for practical applications where predictability is as important as average performance.

\subsubsection{Parameter Adaptation Analysis}
Our online learning framework employs Gaussian Process-based Bayesian optimization to discover optimal Beta distribution parameters for each antibody-antigen system (Table~\ref{tab:adaptation}). 

\begin{table}[h]
\centering
\small  % 9pt
\setlength{\tabcolsep}{3pt} 
\begin{tabular*}{\columnwidth}{@{\extracolsep{\fill}}lccc@{}}
\toprule
Target Type & \multicolumn{2}{c}{Beta Params} & Key Expert \\
& $\alpha$ & $\beta$ & Emphasis \\
\midrule
Small epitopes & 3.2$\pm$0.4 & 1.8$\pm$0.3 & VDW + Recog. \\
(IL17A, IL36R) & & & \\
Large interfaces & 2.1$\pm$0.3 & 3.5$\pm$0.5 & Interface + \\
(ACVR2B, TNFRSF9) & & & Energy \\
Buried epitopes & 2.8$\pm$0.4 & 2.2$\pm$0.3 & Recog. + \\
(FXI) & & & Energy \\
Flexible targets & 2.0$\pm$0.5 & 2.0$\pm$0.5 & All experts \\
(TSLP) & & & balanced \\
\bottomrule
\end{tabular*}
\caption{Target-specific Beta distribution parameters discovered through Bayesian optimization}
\label{tab:adaptation}
\end{table}

\section{Conclusion}

We presented an adaptive physics-guided framework that mimics B cell affinity maturation to optimize antibody design through online learning. This work represents the first application of adaptive guidance to diffusion-based antibody generation, demonstrating that computational design can benefit from mimicking natural immune optimization strategies.

Our key contributions include: (1) formulating antibody design as an adaptive process inspired by B cell affinity maturation, (2) pioneering online optimization for diffusion model guidance without retraining, and (3) developing a multi-expert system that achieves balanced optimization across competing objectives, mirroring natural selection pressures.

The biological significance extends beyond technical metrics. Like B cells iteratively refining antibodies through mutation and selection, our framework continuously learns from each design attempt. This parallel to immune evolution enables rapid adaptation to new antigens—critical for emerging pathogen response and personalized therapeutics. By integrating biological principles with machine learning, we bridge immunology and AI to accelerate therapeutic development.

\section{Limitations and Future Work}
Due to space constraints, a comprehensive discussion of limitations and future research directions is provided in Appendix J.

%\section{Acknowledgments}

\bibliography{aaai2026}
% \clearpage 
% \input{ReproducibilityChecklist}

\clearpage
\appendix
\section* {Appendix A \quad Related Work Extension}
\subsection{Evolution of Protein Design Methods}

\subsubsection{Physics-Based Approaches}
The earliest computational protein design methods were rooted in physical principles, using force fields and energy functions to evaluate and optimize protein structures. Rosetta's design protocols \cite{kuhlman2003design} pioneered the use of Monte Carlo sampling with Metropolis criteria, combining rotamer libraries with knowledge-based potentials. These methods achieved notable successes, including the de novo design of Top7 \cite{kuhlman2003design}, but were computationally expensive and often struggled with conformational sampling in large design spaces.

For antibody design specifically, physics-based approaches faced unique challenges. The hypervariable CDR loops, particularly CDR-H3, exhibited conformational diversity that was difficult to capture with fixed rotamer libraries. Methods like RosettaAntibody \cite{weitzner2017modeling} introduced antibody-specific features such as H3 kink prediction and VH-VL orientation sampling, but remained limited by the accuracy of their underlying energy functions.

\subsubsection{Machine Learning Transition}
The transition to machine learning methods began with the introduction of statistical potentials learned from structural databases. Early work by Yanover et al. \cite{yanover2008extensive} demonstrated that machine-learned potentials could outperform physics-based functions for design tasks. The development of deep learning architectures for protein structure prediction, particularly the success of contact prediction methods \cite{wang2017accurate}, laid the groundwork for learned representations of protein structure.

\subsection{Diffusion Models in Molecular Design}

\subsubsection{Theoretical Foundations}
Diffusion models, originally developed for image generation \cite{ho2020denoising}, have emerged as powerful tools for molecular design due to their ability to model complex distributions and generate diverse samples. The key insight is that molecular structures can be gradually noised and then denoised through a learned reverse process. Song et al. \cite{song2020score} formalized the connection between diffusion models and score-based generative modeling, providing a theoretical framework for continuous-time diffusion processes.

\subsubsection{Applications to Biomolecules}
The adaptation of diffusion models to biomolecular design required addressing the unique geometric constraints of molecular structures. E(3)-equivariant graph neural networks \cite{satorras2021n} enabled models to respect rotational and translational symmetries. GeoDiff \cite{xu2022geodiff} first applied diffusion models to small molecule generation, demonstrating the importance of modeling both atomic positions and types.

For proteins, RFdiffusion \cite{watson2023novo} represented a breakthrough by combining diffusion models with RoseTTAFold's structure prediction capabilities. The model operates directly on backbone coordinates, using SE(3)-equivariant transformers to maintain geometric consistency. Chroma \cite{ingraham2023illuminating} extended this approach with a focus on conditional generation and introduced programmatic control over the design process.

\subsection{Guidance Mechanisms in Generative Models}

\subsubsection{Classifier Guidance}
The concept of guiding diffusion models originated in the image domain with classifier guidance \cite{dhariwal2021diffusion}, where gradients from a pre-trained classifier are used to bias generation toward desired attributes. This approach was extended to classifier-free guidance \cite{ho2022classifier}, eliminating the need for separate classifier training by conditioning the diffusion model itself.

\subsubsection{Energy-Based Guidance}
In molecular design, guidance often takes the form of energy functions or geometric constraints. ProteinMPNN \cite{dauparas2022robust} demonstrated the effectiveness of conditioning on structural features for sequence design. For structure generation, methods have explored various forms of guidance including:

\begin{itemize}
    \item \textbf{Geometric constraints}: SMCDiff \cite{trippe2022diffusion} introduced sequential Monte Carlo sampling to enforce hard constraints during generation.
    \item \textbf{Physics-based potentials}: Several works \cite{jing2024alphafold, krishna2024generalized} have explored incorporating force fields or learned energy functions.
    \item \textbf{Target-specific objectives}: Binding site targeting, as in DiffDock \cite{corso2023diffdock}, uses protein-ligand interaction scores.
\end{itemize}

\subsubsection{Adaptive and Learned Guidance}
Recent work has moved toward adaptive guidance strategies that adjust during the generation process. Anand et al. \cite{anand2022protein} proposed time-dependent guidance schedules for protein design. In the reinforcement learning domain, similar ideas have been explored for molecular optimization \cite{bengio2021flow}, though application to structure generation remains limited.

\subsection{Multi-Agent and Ensemble Approaches}

\subsubsection{Ensemble Methods in Protein Design}
The use of multiple models or objectives in protein design has a rich history. Consensus design approaches \cite{jacobs2016design} aggregate predictions from multiple sequences or structures to identify robust solutions. In the context of machine learning, ensemble methods have been applied to structure prediction \cite{evans2021protein} but less explored for generative design.

\subsubsection{Multi-Objective Optimization}
Protein design inherently involves multiple competing objectives: stability, specificity, solubility, and synthesizability. Traditional approaches used weighted linear combinations or Pareto optimization \cite{nivon2012pareto}. Recent work has explored more sophisticated multi-objective frameworks, including:
\begin{itemize}
    \item Differentiable multi-objective optimization \cite{jain2022biological}
    \item Reinforcement learning with multiple rewards \cite{wu2024practical}
    \item Bayesian optimization for expensive objectives \cite{khan2024bayesian}
\end{itemize}

\subsection{Antibody-Specific Considerations}

\subsubsection{CDR Modeling Challenges}
The complementarity-determining regions (CDRs) of antibodies present unique modeling challenges due to their structural diversity and functional importance. Traditional template-based methods \cite{lepore2017pigs} relied on canonical structure classification but struggled with non-canonical conformations. Machine learning approaches have progressively improved CDR modeling:

\begin{itemize}
    \item \textbf{Sequence-based prediction}: Early neural networks \cite{ruffolo2020geometric} predicted CDR structures from sequence alone.
    \item \textbf{Context-aware modeling}: Methods like AbodyBuilder \cite{leem2016abodybuilder} incorporated framework context.
    \item \textbf{End-to-end learning}: Recent approaches \cite{ruffolo2023igfold} jointly model all CDRs and the framework.
\end{itemize}

\subsubsection{Antibody-Antigen Interface Design}
Designing antibodies that bind specific epitopes requires modeling complex protein-protein interactions. Computational approaches have evolved from docking-based methods \cite{gray2003protein} to machine learning frameworks:

\begin{itemize}
    \item \textbf{Epitope-focused design}: Methods targeting specific surface patches \cite{adolf2018affinity}
    \item \textbf{Affinity maturation simulation}: Computational mimicry of somatic hypermutation \cite{tiller2017advances}
    \item \textbf{Developability optimization}: Incorporating pharmaceutical properties \cite{raybould2019five}
\end{itemize}

% \subsection{Evaluation Metrics and Benchmarks}

% \subsubsection{Structural Quality Assessment}
% Evaluating generated protein structures requires multiple complementary metrics:

% \begin{itemize}
%     \item \textbf{Geometric validity}: Ramachandran outliers, clash scores, and bond geometry \cite{williams2018molprobity}
%     \item \textbf{Energetic assessment}: Rosetta energy, FoldX stability predictions \cite{schymkowitz2005foldx}
%     \item \textbf{Designability}: Sequence recovery and structural consistency \cite{watson2023novo}
% \end{itemize}

% \subsubsection{Antibody-Specific Metrics}
% Antibody evaluation requires specialized metrics beyond general protein quality:

% \begin{itemize}
%     \item \textbf{Humanness scores}: Assessing immunogenicity risk \cite{marks2021humanization}
%     \item \textbf{Developability indices}: Aggregation propensity, stability \cite{jain2017biophysical}
%     \item \textbf{Structural antibody numbering}: Consistency with known frameworks \cite{dunbar2016sabdab}
% \end{itemize}

\subsection{Future Directions and Open Challenges}

Despite significant progress, several challenges remain in computational antibody design:

\begin{itemize}
    \item \textbf{Functional validation gap}: Computational metrics poorly predict experimental binding affinity
    \item \textbf{Multi-specificity design}: Engineering antibodies that bind multiple targets
    \item \textbf{Beyond structure}: Incorporating dynamics and conformational ensembles
    \item \textbf{Manufacturing considerations}: Designing for expression yield and stability
\end{itemize}

The integration of physics-based insights with machine learning, as demonstrated in our approach, represents a promising direction for addressing these challenges. Future work may benefit from:

\begin{itemize}
    \item Active learning loops with experimental validation
    \item Incorporation of molecular dynamics simulations
    \item Multi-scale modeling from sequence to therapeutic properties
    \item Integration with automated laboratory systems
\end{itemize}

\subsection{Biophysical Basis of Multi-Expert System}

\subsubsection{Van der Waals Forces}
Van der Waals interactions are ubiquitous in protein-protein interfaces and consist of both attractive dispersion forces and repulsive steric clashes. In antibody-antigen recognition, these forces govern the shape complementarity and close packing of interface residues:

\begin{equation*}
V_{\text{vdW}}(r) = 4\epsilon \left[ \left(\frac{\sigma}{r}\right)^{12} - \left(\frac{\sigma}{r}\right)^{6} \right]
\end{equation*}

where $\epsilon$ represents the depth of the potential well, $\sigma$ is the finite distance at which the potential is zero, and $r$ is the distance between atoms. The $r^{-12}$ term represents short-range repulsion due to Pauli exclusion, while the $r^{-6}$ term captures London dispersion forces.

\subsubsection{Electrostatic Interactions}
Electrostatic forces play a crucial role in initial recognition and orientation of antibody-antigen complexes. These long-range forces guide the approach of binding partners:

\begin{equation*}
V_{\text{elec}}(r) = \frac{q_1 q_2}{4\pi\epsilon_0\epsilon_r r}
\end{equation*}

where $q_1$ and $q_2$ are the charges, $\epsilon_0$ is the vacuum permittivity, and $\epsilon_r$ is the relative permittivity of the medium. In aqueous environments, salt bridges and hydrogen bonds provide specificity and stability to the interface.

\subsubsection{Hydrophobic Effects}
The hydrophobic effect drives the burial of nonpolar surface area upon binding, contributing significantly to binding affinity:

\begin{equation*}
\Delta G_{\text{hydrophobic}} = \gamma \cdot \Delta A_{\text{buried}}
\end{equation*}

where $\gamma$ is the surface tension coefficient (typically 20-30 cal/mol/\AA$^2$) and $\Delta A_{\text{buried}}$ is the change in solvent-accessible surface area.

\subsubsection{Conformational Entropy}
The entropic cost of restricting conformational flexibility upon binding must be overcome by favorable enthalpic interactions:

\begin{equation*}
\Delta S_{\text{conf}} = -R \sum_i p_i \ln p_i
\end{equation*}

where $p_i$ represents the probability of each conformational state.

\subsection{Multi-Physics Coupling in Molecular Dynamics}

The precedent for multi-physics coupling in molecular simulations is well-established:

\subsubsection{Force Field Integration}
Modern molecular dynamics force fields (e.g., AMBER, CHARMM, OPLS) inherently combine multiple physical terms:

\begin{equation*}
V_{\text{total}} = V_{\text{bond}} + V_{\text{angle}} + V_{\text{dihedral}} + V_{\text{vdW}} + V_{\text{elec}}
\end{equation*}

Each term is computed simultaneously at every timestep, demonstrating that multiple physical forces can be successfully integrated without conflict.

\subsubsection{Enhanced Sampling Methods}
Methods like metadynamics and umbrella sampling apply multiple biasing potentials simultaneously:

\begin{equation*}
V_{\text{bias}}(s,t) = \sum_i w_i \exp\left(-\frac{(s-s_i(t))^2}{2\sigma^2}\right)
\end{equation*}

These approaches prove that multiple guiding forces can work synergistically to explore complex energy landscapes.

\subsection{Natural Precedents in Antibody Maturation}

The somatic hypermutation and affinity maturation process in B cells provides a biological precedent for multi-objective optimization:

\begin{itemize}
    \item \textbf{Selection Pressure 1}: Binding affinity to antigen
    \item \textbf{Selection Pressure 2}: Stability and expressibility
    \item \textbf{Selection Pressure 3}: Avoidance of self-reactivity
    \item \textbf{Selection Pressure 4}: Resistance to proteolysis
\end{itemize}

B cells that successfully balance all these pressures survive and proliferate, demonstrating that nature itself employs multi-criteria optimization in antibody development.

\subsection{Mathematical Framework for Conflict Resolution}

When multiple experts suggest different gradients, the combined effect can be understood through vector decomposition:

\begin{equation*}
\mathbf{g}_{\text{total}} = \sum_i w_i \mathbf{g}_i = \mathbf{g}_{\parallel} + \mathbf{g}_{\perp}
\end{equation*}

where $\mathbf{g}_{\parallel}$ represents aligned components (constructive) and $\mathbf{g}_{\perp}$ represents orthogonal components (non-conflicting). In 3D space with $N$ atoms, the probability of direct opposition decreases as:

\begin{equation*}
P(\text{conflict}) \propto \frac{1}{3N}
\end{equation*}

\section* {Appendix B \quad Derivation of the Score Function on the SO(3) Manifold}

\subsection{IGSO3 Distribution}
The IGSO$_3$ density on the manifold SO(3), centered at $\mathbf{R}_0$ with noise parameter~$\sigma_t$, is

\begin{equation*}
f\bigl(\mathbf{R}_t \mid \mathbf{R}_0, \sigma_t\bigr)
= \frac{1}{Z(\sigma_t)}
  \exp\!\biggl(\frac{\mathrm{tr}\bigl(\mathbf{R}_0^T \mathbf{R}_t\bigr)}{\sigma_t^2}\biggr),
\end{equation*}

where the normalization constant $Z(\sigma_t)$ does not depend on $\mathbf{R}_t$. Taking the logarithm gives

\begin{equation*}
\log q\bigl(\mathbf{R}_t \mid \mathbf{R}_0,\sigma_t\bigr)
= \frac{\mathrm{tr}\bigl(\mathbf{R}_0^T \mathbf{R}_t\bigr)}{\sigma_t^2}
  - \log Z(\sigma_t),
\end{equation*}
so that the constant term drops out under differentiation. Taking the derivative of the trace term with respect to~$\mathbf{R}_t$ yields the Euclidean gradient

\begin{equation*}
\frac{\partial}{\partial \mathbf{R}_t}\,\log q
= \frac{1}{\sigma_t^2}\,\mathbf{R}_0.
\end{equation*}

The tangent space at $\mathbf{R}_t\in\mathrm{SO}(3)$ consists of matrices of the form $\mathbf{R}_t\,\Omega$ with $\Omega^T=-\Omega$, and the orthogonal projection of any matrix $X$ onto this tangent space is

\begin{equation*}
P_{\mathbf{R}_t}(X)
= \mathbf{R}_t \,\mathrm{skew}\!\bigl(\mathbf{R}_t^T X\bigr),
\:
\mathrm{skew}(A) = \tfrac12\,(A - A^T).
\end{equation*}

Projecting the Euclidean gradient onto $T_{\mathbf{R}_t}\mathrm{SO}(3)$ gives the Riemannian gradient:

\begin{equation*}
\nabla_{\mathbf{R}_t}\log q
= P_{\mathbf{R}_t}\Bigl(\tfrac{1}{\sigma_t^2}\,\mathbf{R}_0\Bigr)
= \frac{1}{\sigma_t^2} \mathbf{R}_t\,\mathrm{skew}\!\bigl(\mathbf{R}_t^T \mathbf{R}_0\bigr).
\end{equation*}

Thus the score function on SO(3) is

\begin{equation*}
\mathbf{s}(\mathbf{R}_t, t)
= \nabla_{\mathbf{R}_t}\log q
= \frac{1}{\sigma_t^2}\,\mathbf{R}_t\,\mathrm{skew}\!\bigl(\mathbf{R}_t^T \mathbf{R}_0\bigr).
\end{equation*}

\subsection{Reverse Diffusion}
The reverse diffusion process uses the score to denoise:
\begin{equation*}
\mathbf{R}_{t-\Delta t} = \mathbf{R}_t \exp\left(\frac{\Delta t}{2\sigma_t^2} \text{skew}(\mathbf{R}_t^T\mathbf{R}_0) + \sqrt{\Delta t} \sigma_t \boldsymbol{\xi}\right)
\end{equation*}
where $\boldsymbol{\xi} \sim \mathcal{N}(0, \mathbf{I}_3)$.

\section* {Appendix C \quad Information-Theoretic Derivation for Adaptive Guidance}

Consider the forward diffusion process
\[
\mathbf x_t = \sqrt{\alpha_t}\,\mathbf x_0 + \sqrt{1-\alpha_t}\,\boldsymbol\varepsilon,
\quad
\boldsymbol\varepsilon\sim\mathcal N(\mathbf0,\mathbf I_d),
\quad
\sigma_t^2 = 1 - \alpha_t.
\]
Define the instantaneous signal‐to‐noise ratio
\[
\mathrm{SNR}(t)
= \frac{\alpha_t}{\sigma_t^2}
= \frac{\alpha_t}{1-\alpha_t}.
\]
Then the mutual information
\[
I_t \;=\; I(\mathbf x_t;\mathbf x_0)
\]
satisfies the following properties:
\begin{itemize}
  \item \textbf{Monotonicity in SNR.} Reparameterize the channel as
  \(\mathbf y = \sqrt{\gamma}\,\mathbf x_0 + \mathbf n\),
  with \(\gamma = \mathrm{SNR}(t)\) and \(\mathbf n\sim\mathcal N(\mathbf0,\mathbf I_d)\).
  By the Guo–Shamai–Verdú identity,
  \[
    \frac{d}{d\gamma}\,I(\gamma)
    = \tfrac12\,\mathrm{mmse}(\gamma)
    > 0,
  \]
  where 
  \(\mathrm{mmse}(\gamma)
    = \mathbb E\|\mathbf x_0 - \mathbb E[\mathbf x_0\mid\mathbf y]\|^2\).
  This directly shows that as 
  \(\gamma = \frac{\alpha_t}{1-\alpha_t}\) increases, \(I_t\) must also increase.

  \item \textbf{Low‐SNR approximation.} As \(\gamma\to 0\), 
  \(\mathrm{mmse}(\gamma)\to \mathrm{Var}(\mathbf x_0) = d\)
  (assuming unit variance per coordinate).  Thus
  \[
    I_t
    = \int_0^\gamma \tfrac12\,\mathrm{mmse}(\gamma')\,d\gamma'
    \approx \int_0^\gamma \tfrac12\,d\,d\gamma'
    = \frac{d}{2}\,\gamma
    = \frac{d}{2}\,\mathrm{SNR}(t).
  \]

  \item \textbf{Capacity bound.}
  Under the second‐moment constraint \(\mathbb E\|\mathbf x_0\|^2/d=1\),
  the AWGN capacity bound gives
  \[
    I_t \le \frac{d}{2}\,\log\bigl(1+\mathrm{SNR}(t)\bigr)
    = \frac{d}{2}\,\log\!\Bigl(1+\frac{\alpha_t}{\sigma_t^2}\Bigr),
  \]
  confirming the same low‐SNR slope and logarithmic high‐SNR scaling.
\end{itemize}
In particular, one may write
\[
I_t = g\!\bigl(\mathrm{SNR}(t)\bigr),
\]
for some strictly increasing \(g:\mathbb R^+\to\mathbb R^+\), and in the low‐SNR regime
\[
I_t \approx \frac{d}{2}\,\frac{\alpha_t}{\sigma_t^2}.
\]

\section*{Appendix D \quad Signal-to-Noise Ratio and Feature Identifiability}

Building on the information-theoretic framework in Appendix C, we analyze how different structural features become identifiable at various noise levels during the diffusion process.

\subsection*{Feature Emergence Hierarchy}

From the relationship $I_t \approx \frac{d}{2}\frac{\alpha_t}{\sigma_t^2}$, we can derive when different structural features contain sufficient information for reliable guidance. Let $I_{\text{min}}^{(f)}$ denote the minimum mutual information required to identify feature $f$.

\textbf{Global Structure (High Noise, Low SNR):}
At early timesteps where SNR $< 0.1$:
- Coarse-grained features like overall protein fold and domain arrangement
- Approximate center of mass positions
- Large-scale charge distribution

These features require minimal information ($I_{\text{min}}^{\text{global}} \approx 0.1 \cdot d$) and emerge first.

\textbf{Interface Geometry (Medium Noise, Medium SNR):}
When $0.1 < \text{SNR} < 1.0$:
- Binding interface orientation
- Approximate contact surface area
- CDR loop topology

These features require moderate information ($I_{\text{min}}^{\text{interface}} \approx 0.3 \cdot d$).

\textbf{Atomic Details (Low Noise, High SNR):}
For SNR $> 1.0$:
- Specific atomic contacts
- Hydrogen bonding patterns
- Sidechain conformations

These fine-grained features require high information ($I_{\text{min}}^{\text{atomic}} \approx 0.6 \cdot d$).

\subsection*{Expert Activation Timing}

This hierarchy justifies our adaptive expert routing:
\begin{itemize}
\item \textbf{Interface Quality Expert}: Most effective at high noise levels (SNR $< 0.5$) when global geometric features are identifiable but atomic details are obscured.

\item \textbf{Energy Balance Expert}: Operates optimally at medium SNR (0.3-1.5) when contact surfaces are defined but specific interactions remain flexible.

\item \textbf{Molecular Recognition Expert}: Requires moderate clarity (SNR $> 0.5$) to identify hotspot-CDR relationships.

\item \textbf{VDW Balance Expert}: Most critical at low noise (SNR $> 1.0$) when atomic positions are sufficiently resolved to detect clashes.
\end{itemize}

\subsection*{Information-Guided Scheduling}

The temporal modulation function (Eq. 11) can be interpreted through this lens:
\begin{equation*}
\lambda_i(t) \propto \frac{\partial I_t}{\partial \text{SNR}} \cdot \text{FeatureSensitivity}_i(\text{SNR}(t))
\end{equation*}

where $\text{FeatureSensitivity}_i$ peaks when the SNR range matches the information requirements of expert $i$'s target features. This ensures each expert is maximally active when the diffusion process contains appropriate information for its guidance, avoiding both premature intervention (insufficient information) and delayed action (features already fixed).

\section*{Appendix E \quad Temporal Guidance Profiles}

\subsection*{Beta Distribution Temporal Modulation}

The temporal modulation function $f_{\text{temporal}}(t, \alpha, \beta)$ controls expert activation strength throughout the diffusion process. The Beta distribution provides flexible shaping:

\begin{equation*}
f_{\text{temporal}}(t, \alpha, \beta) = \frac{B(t_{\text{norm}}; \alpha, \beta)}{B(t_{\text{mode}}; \alpha, \beta)} \cdot \lambda_{\text{peak}}
\end{equation*}

where $t_{\text{mode}} = \frac{\alpha - 1}{\alpha + \beta - 2}$ for $\alpha, \beta > 1$, and $t_{\text{norm}} = \frac{t-1}{T-1}$ maps timesteps to [0,1].

\subsection*{Characteristic Profiles}

Different $(\alpha, \beta)$ configurations produce distinct activation patterns. We give some examples here:

\textbf{Early-peaking profiles} ($\alpha < \beta$):
\begin{itemize}
\item $(\alpha=2, \beta=5)$: Sharp early peak at $t_{\text{mode}} = 0.2$
\item Used for Interface Quality Expert to establish global structure
\item Gradually decreases to allow fine-tuning by other experts
\end{itemize}

\textbf{Late-peaking profiles} ($\alpha > \beta$):
\begin{itemize}
\item $(\alpha=5, \beta=2)$: Peak at $t_{\text{mode}} = 0.8$
\item Applied to VDW Balance Expert for atomic clash resolution
\item Minimal influence during early structure formation
\end{itemize}

\textbf{Balanced profiles} ($\alpha \approx \beta$):
\begin{itemize}
\item $(\alpha=3, \beta=3)$: Symmetric bell curve centered at $t_{\text{mode}} = 0.5$
\item Suitable for Energy Balance and Molecular Recognition Experts
\item Consistent influence throughout generation
\end{itemize}

\subsection*{Adaptive Strength Modulation}

The final guidance strength combines base strength, temporal modulation, and adaptive weighting:

\begin{equation*}
\text{Guidance}_i(t) = \lambda_{\text{base},i} \cdot f_{\text{temporal}}(t, \alpha_i, \beta_i) \cdot w_i(t)
\end{equation*}

where:
\begin{itemize}
\item $\lambda_{\text{base},i}$: Expert-specific base strength (VDW: 0.5, Recognition: 2.0, Energy: adaptive, Interface: 1.0)
\item $f_{\text{temporal}}$: Beta distribution modulation $\in [0, \lambda_{\text{peak}}]$
\item $w_i(t)$: Problem-driven adaptive weight from expert router
\end{itemize}

This three-factor modulation ensures appropriate guidance strength based on expert type, diffusion progress, and current structural problems.

\section*{Appendix F \quad Architecture and Data Flow of the Adaptive Multi-Expert Diffusion Framework}

The adaptive framework takes as input an antibody framework PDB file and a target antigen structure, along with hotspot residues and CDR length specifications. During generation, RFdiffusion produces noisy backbone coordinates at each timestep, which are monitored by our real-time metrics module to compute clash severity, hotspot coverage, contact density, and interface geometry scores. These metrics activate relevant expert modules that output guidance gradients in $\mathbb{R}^{N \times 3}$, which are combined with adaptive weights determined by learned Beta distribution parameters. The guided diffusion process outputs designed antibody backbones, which are then processed by ProteinMPNN for sequence design and RoseTTAFold2 for validation, producing final antibody-antigen complexes with associated confidence metrics (pAE, ipAE, pLDDT). After each batch of designs, the Bayesian optimization module updates the guidance parameters based on aggregated performance metrics, creating a feedback loop that continuously improves the guidance strategy for each specific target.

\section*{Appendix G \quad Hotspot Definition and Selection Methodology}

\subsection*{Hotspot Definition}

Epitope hotspots are critical residues on the antigen surface that contribute disproportionately to the binding energy of antibody-antigen interactions. Unlike general contact residues that may only provide marginal stability, hotspots typically contribute more than 2 kcal/mol to the binding free energy and their mutation to alanine results in significant loss of affinity. In antibody-antigen recognition, hotspots serve as anchoring points that nucleate the binding interface and often determine specificity. Studies have shown that while a typical antibody-antigen interface may contain 15-20 contact residues, only 3-5 of these are true hotspots that dominate the energetics. This concentration of binding energy in a few key residues allows antibodies to achieve high affinity while maintaining the flexibility needed for induced fit binding.

\subsection*{Hotspot Identification Criteria}

We employ multiple complementary criteria to identify hotspot residues:

\textbf{Distance-based criterion:} A residue is considered a potential hotspot if the average C$\beta$ distance to the five nearest antibody CDR residues is less than 8 \r{A}. This metric captures residues that are deeply buried in the binding interface and likely to form multiple interactions.

\textbf{Energetic contribution:} Using computational alanine scanning, we identify residues where $\Delta\Delta G_{\text{binding}} > 2$ kcal/mol. This threshold distinguishes energetically important residues from those making only peripheral contacts.

\textbf{Buried surface area (BSA):} Hotspot candidates must bury at least 40 \r{A}$^2$ of solvent-accessible surface area upon complex formation. This ensures the residue forms substantial contacts rather than incidental interactions.

\textbf{Evolutionary conservation:} We prioritize residues with conservation scores above 0.7 (normalized scale 0-1) across homologous proteins, as evolutionary pressure often preserves functionally critical residues. Conservation is calculated using ConSurf \cite{ashkenazy2016consurf} with default parameters on alignments of at least 50 homologous sequences.

For our experiments, we selected 5 hotspot residues per target by ranking all interface residues according to a combined score: $S = 0.4 \cdot S_{\text{energy}} + 0.3 \cdot S_{\text{BSA}} + 0.2 \cdot S_{\text{distance}} + 0.1 \cdot S_{\text{conservation}}$, where each component score is normalized to [0,1].
\section*{Appendix H \quad Adaptive Expert Routing and Severity Score Functions}

\subsection*{Problem Severity Score Functions}

The severity functions $f_i(\text{metrics}_t) \in [0, 1]$ evaluate the urgency of different structural issues. Each function incorporates problem-specific scaling factors to balance their relative importance:

\textbf{VDW Balance Expert:} Activated by steric clashes, with severity proportional to the violation depth:
\begin{equation*}
f_{\text{vdw}} \propto \max\left(0, \frac{r_{\text{clash}} - d_{\text{min}}}{r_{\text{clash}}}\right)
\end{equation*}

\textbf{Molecular Recognition Expert:} Responds to inadequate hotspot coverage:
\begin{equation*}
f_{\text{hotspot}} \propto \frac{|\mathcal{H}_{\text{uncovered}}|}{|\mathcal{H}|}
\end{equation*}

\textbf{Energy Balance Expert:} Addresses suboptimal contact density:
\begin{equation*}
f_{\text{contact}} \propto \begin{cases}
|n_c - n_{\text{target}}| / n_{\text{target}} & \text{if outside optimal range} \\
0 & \text{otherwise}
\end{cases}
\end{equation*}

\textbf{Interface Quality Expert:} Evaluates geometric properties through a weighted combination of planarity, uniformity, and cavity metrics.

\subsection*{System-Specific Threshold Adaptation}

The specific scaling factors and activation thresholds should be adapted based on the unique characteristics of each antibody-antigen system. This adaptation follows the principle of system-specific optimization established in therapeutic antibody engineering (Raybould et al. 2019), where different epitope types—exposed surfaces, buried pockets, or conformational epitopes—require distinct optimization strategies.

For instance, targets with shallow epitopes may require relaxed contact density thresholds ($n_{\text{target}} \in [8, 25]$), while deep pocket binding sites benefit from stricter clash penalties and tighter contact requirements ($n_{\text{target}} \in [15, 35]$). The activation threshold $\theta_{\text{min}}$ should similarly be adjusted based on the inherent flexibility and binding mode of the target system, ensuring that expert guidance remains appropriate for the specific molecular recognition challenge.

This system-specific calibration is theoretically grounded in the diversity of antibody-antigen recognition modes observed in structural databases (Choi et al. 2018), where optimal interface properties vary significantly across different antigen classes. The framework's online learning component (Section 5) automatically discovers these system-specific parameters through iterative optimization, eliminating the need for manual tuning while ensuring biologically relevant guidance.

\section*{Appendix I \quad Convergence Analysis and Computational Requirements}

\subsection*{Convergence Analysis}

Our Bayesian optimization framework exhibits a two-stage convergence behavior. Initially, the system explores broadly to identify promising parameter regions, typically discovering high-performing configurations within 100 design evaluations. As online learning progresses, the framework progressively refines these regions, converging to increasingly precise parameter values.

The Expected Improvement (EI) acquisition function guides this exploration-exploitation trade-off:
\begin{equation*}
\text{EI}(\boldsymbol{\theta}) = (f^* - \mu(\boldsymbol{\theta}))\Phi\left(\frac{f^* - \mu(\boldsymbol{\theta})}{\sigma(\boldsymbol{\theta})}\right) + \sigma(\boldsymbol{\theta})\phi\left(\frac{f^* - \mu(\boldsymbol{\theta})}{\sigma(\boldsymbol{\theta})}\right)
\end{equation*}

where $\Phi$ and $\phi$ are the CDF and PDF of the standard normal distribution, and $\mu(\boldsymbol{\theta})$, $\sigma(\boldsymbol{\theta})$ are the GP posterior mean and standard deviation.

\textbf{Convergence Behavior:}
\begin{itemize}
\item \textbf{Iterations 1-100}: Broad exploration phase. The GP posterior uncertainty is high, leading to diverse parameter sampling. Success rate improves from baseline 36.7\% to approximately 39-40\%.
\item \textbf{Iterations 100-500}: Refinement phase. The framework identifies optimal parameter regions (typically $\alpha \in [1.5, 3.5]$, $\beta \in [1.5, 4.0]$ for most targets) and begins focused exploitation.
\item \textbf{Iterations 500+}: Fine-tuning phase. Parameter uncertainty reduces to $\sigma_{\alpha}, \sigma_{\beta} < 0.2$, achieving stable performance with success rates exceeding 41\%.
\end{itemize}

This progressive refinement mimics the natural antibody maturation process, where initial broad diversity gradually converges to high-affinity variants through iterative selection.

\subsection*{Computational Requirements}

The adaptive guidance framework introduces modest computational overhead compared to baseline RFAntibody:

\begin{center}
\begin{tabular}{lcc}
\hline
\textbf{System Size} & \textbf{Baseline} & \textbf{Ours} \\
\hline
Small ($<$ 200 res.) & 4.8 ± 0.8 min & 5.5 ± 0.9 min \\
Medium (200-400) & 8.2 ± 1.2 min & 9.4 ± 1.4 min \\
Large ($>$ 400 res.) & 13.1 ± 2.1 min & 15.0 ± 2.4 min \\
\hline
\textbf{Average overhead} & -- & \textbf{+14.6\%} \\
\hline
\end{tabular}
\end{center}

\textit{Hardware: 4× NVIDIA RTX A6000 GPUs (48 GB VRAM each)}

The computational overhead of approximately 15\% is offset by substantial improvements in design quality. The success rate increase from 36.7\% to 41.1\% means fewer designs need to be generated to identify viable candidates, ultimately reducing the total computational cost for successful antibody discovery campaigns.

\subsection*{Parameter Aggregation Strategy}

To handle stochastic evaluation noise, we aggregate results from parameter configurations within a neighborhood radius $r = 0.5$ in the normalized parameter space. This reduces variance in the GP training data while maintaining sufficient resolution for optimization. Promising configurations (top 10\% by EI) are re-evaluated every 50 iterations to refine uncertainty estimates.

\section* {Appendix J \quad Equivariance Preservation in Guided Diffusion}

\subsection{Theoretical Foundation}

The work on Unified Guidance for Geometry-Conditioned Molecular Generation \cite{guan2023unified} establishes important principles for maintaining SE(3)-equivariance during guided generation.

\subsection{SE(3)-Equivariance in Our Framework}

\subsubsection{Equivariant Gradient Design}
Each expert in our system computes gradients that respect rotational and translational symmetries:

\begin{itemize}
    \item \textbf{VDW Balance Expert}: Gradients computed from pairwise distances are inherently SE(3)-equivariant
    \item \textbf{Molecular Recognition Expert}: Hotspot-CDR interactions use only relative positions
    \item \textbf{Energy Balance Expert}: Contact density calculations depend on distance matrices
    \item \textbf{Interface Quality Expert}: PCA-based analysis operates on centered coordinates
\end{itemize}

The key insight is that all guidance gradients are functions of inter-atomic distances and relative orientations, never absolute positions. This ensures:
\begin{equation*}
    \mathbf{g}(R\mathbf{x} + \mathbf{t}) = R\mathbf{g}(\mathbf{x})
\end{equation*}
where $\mathbf{g}$ represents the guidance gradient, $R \in SO(3)$ is a rotation matrix, and $\mathbf{t} \in \mathbb{R}^3$ is a translation vector.

\subsubsection{Preserving Diffusion Model Properties}
RFdiffusion's SE(3)-equivariant architecture relies on:
\begin{itemize}
    \item Invariant node features (distances, angles)
    \item Equivariant edge updates using geometric vectors
    \item Frame-based coordinate systems for local geometry
\end{itemize}

Our guidance mechanism operates in the same coordinate space as the diffusion model, applying corrections that maintain these properties. Specifically:

\begin{enumerate}
    \item Gradients are computed in the model's working coordinates
    \item No absolute position references are introduced
    \item All geometric calculations use invariant features
    \item The guidance strength modulation is scalar (invariant)
\end{enumerate}

\section*{Limitations and Failure Analysis}

\subsection*{Common Failure Modes}

Despite overall improvements, our adaptive guidance framework exhibits specific failure patterns that provide insights for future development:

\textbf{Shallow Epitope Challenges:} For targets with flat, featureless surfaces (e.g., certain viral proteins), the framework struggles to establish stable binding modes. The lack of geometric features provides insufficient signal for the Interface Quality Expert, resulting in success rates dropping to ~25\% (compared to the 41\% average) and a higher reliance on CDR-H3 length variation.

\textbf{Temporal Misalignment:} The Beta-distribution parameterization occasionally produces suboptimal activation schedules, particularly for targets requiring early atomic precision (e.g., zinc-finger domains), systems with multiple binding modes of similar energy, and cases where the optimal solution violates typical antibody-binding patterns.

\subsection*{Computational Limitations}

\textbf{Scalability Constraints:} While the 15\% computational overhead is acceptable for research applications, production-scale campaigns generating $>$10,000 designs face challenges:
\begin{itemize}
\item Bayesian optimization convergence slows beyond 1,000 iterations
\item GPU memory requirements scale linearly with batch size
\item Expert gradient computation becomes the bottleneck for very large systems ($>$600 residues)
\end{itemize}

\textbf{Hyperparameter Sensitivity:} The framework's performance depends on several hyperparameters that currently require manual setting:
\begin{itemize}
\item Base guidance strengths ($\lambda_{\text{base},i}$)
\item Activation threshold ($\theta_{\text{min}}$)
\item GP kernel parameters
\end{itemize}

\subsubsection{Current Limitations}
Despite significant improvements, several limitations remain:

\begin{enumerate}
    \item \textbf{Sequence-Structure Coupling}: The current approach optimizes structure before sequence, potentially missing co-evolutionary patterns
    \item \textbf{Dynamic Considerations}: Static structure optimization may not capture important binding dynamics
    \item \textbf{Evaluation Metrics}: Computational metrics imperfectly predict experimental outcomes
    \item \textbf{Computational Cost}: Full pipeline evaluation remains expensive for large-scale screening
\end{enumerate}

\subsubsection{Proposed Improvements}
Future work should address these limitations through:

\begin{itemize}
    \item \textbf{Joint Optimization}: Simultaneous sequence-structure optimization using coupled diffusion processes
    \item \textbf{Ensemble Methods}: Generating and evaluating conformational ensembles rather than single structures
    \item \textbf{Learned Metrics}: Training evaluation networks on experimental validation data
    \item \textbf{Efficient Surrogates}: Developing faster approximations for expensive evaluation steps
\end{itemize}

\subsection{Broader Impact Analysis}

\subsubsection{Therapeutic Development}
The improved success rate of our approach has immediate implications for therapeutic antibody development:

\begin{itemize}
    \item Reduced experimental screening requirements lower development costs
    \item Faster iteration cycles accelerate time-to-clinic for novel therapeutics
    \item Enhanced diversity of generated designs expands the therapeutic antibody space
    \item Improved hotspot targeting enables addressing previously intractable targets
\end{itemize}

\subsubsection{Methodological Contributions}
Beyond antibody design, our approach contributes several generalizable innovations:
\begin{itemize}
    \item The multi-expert guidance framework applies to any multi-objective molecular design problem
    \item Bayesian optimization-based hyperparameter adaptation can optimize other generative models
    \item The routing system provides a template for dynamic algorithm selection
    \item Batch evaluation strategies apply to any expensive optimization problem
\end{itemize}

\subsection{Conclusion}
This analysis demonstrates that our enhanced antibody design system successfully addresses key limitations of existing approaches through innovative guidance mechanisms and adaptive optimization. The combination of physics-informed expertise, intelligent routing, and continuous learning creates a robust framework for practical antibody design. While limitations remain, the significant improvements in both computational metrics and experimental validation rates validate our approach and suggest promising directions for future research.

\section{Future Work}

\subsection{Extended Diffusion Models for Protein Design}

\subsubsection{Geometry-Conditioned Generation}
Recent advances in equivariant neural networks open new possibilities for protein design. The work on \textit{Unified Guidance for Geometry-Conditioned Molecular Generation} demonstrates how SE(3)-equivariant architectures can better preserve structural constraints during generation. Future work should explore:

\begin{itemize}
    \item Integration of higher-order geometric features beyond pairwise distances
    \item Conditional generation with partial structure constraints
    \item Multi-scale equivariant representations capturing both local and global geometry
\end{itemize}

\subsubsection{Noise Schedule Optimization}
Our preliminary experiments suggest that reducing generation noise to 0.01 significantly improves convergence prediction. This observation motivates several research directions:

\begin{itemize}
    \item Adaptive noise schedules that adjust based on generation quality
    \item Target-specific noise calibration using early trajectory analysis
    \item Theoretical analysis of the noise-convergence relationship in protein diffusion models
\end{itemize}

\subsection{Advanced Guidance Methods}

\subsubsection{Meta-Learning for Guidance Selection}
Different antibody types require fundamentally different optimization strategies. A meta-learning framework could:

\begin{itemize}
    \item Learn to predict optimal reward thresholds from antibody class features
    \item Adapt guidance strategies based on limited initial samples
    \item Transfer knowledge across similar antibody design tasks
    \item Reduce the warm-up phase from multiple rounds to single-shot prediction
\end{itemize}

\subsubsection{Multi-Armed Bandit Formulations}
The current Bayesian optimization approach could be extended using more sophisticated bandit algorithms:
\begin{itemize}
    \item Contextual bandits incorporating antigen features
    \item Non-stationary bandits for time-varying rewards
    \item Hierarchical bandits for multi-level decision making
    \item Thompson sampling for better exploration-exploitation balance
\end{itemize}
\subsection{Online Learning Enhancements}

\subsubsection{Delayed Feedback Handling}
The computational protein design pipeline inherently involves delayed feedback: design → folding → evaluation. Future work should address:

\begin{itemize}
    \item Surrogate models for early performance prediction
    \item Asynchronous update schemes for parallel evaluation
    \item Credit assignment methods for long evaluation chains
    \item Variance reduction techniques for noisy feedback
\end{itemize}

\subsubsection{Reward Engineering}
Different reward formulations dramatically affect convergence speed and final quality:

\begin{itemize}
    \item Multi-objective reward learning from experimental preferences
    \item Automated reward shaping based on failure mode analysis
    \item Curriculum learning with progressively complex reward functions
    \item Inverse reinforcement learning from successful designs
\end{itemize}

\subsection{Antibody-Specific Advances}

\subsubsection{Antibody Class Specialization}
Future systems should account for the unique characteristics of different antibody classes:

\begin{itemize}
    \item IgG-specific optimization for therapeutic applications
    \item Nanobody design with single-domain constraints
    \item Bispecific antibody generation with dual-target optimization
    \item Antibody-drug conjugate compatibility scoring
\end{itemize}

\subsubsection{Epitope-Aware Generation}
Moving beyond hotspot targeting to full epitope modeling:

\begin{itemize}
    \item Conformational epitope prediction and targeting
    \item Multi-state epitope recognition for viral escape prevention
    \item Cryptic epitope accessibility optimization
    \item Epitope-paratope co-evolution modeling
\end{itemize}

\section{Discussion}

\subsection{Balanced Metric Optimization}

Our work fundamentally addresses the challenge of balanced metric optimization in antibody design. Traditional approaches often excel in one dimension (e.g., binding affinity) while failing in others (e.g., developability). Our multi-expert framework ensures that all critical metrics receive appropriate attention throughout the generation process.

The key insight is that different metrics require optimization at different timescales. Structural validity must be established early, while fine-grained energetic optimization can occur later. This temporal hierarchy, discovered through our adaptive learning approach, appears to be a general principle that could extend beyond antibody design.

\subsection{Theoretical Justification}

\subsubsection{Biological Motivation}
Our adaptive guidance framework is grounded in several biological observations:

\begin{itemize}
    \item Different antigens present unique structural challenges requiring tailored approaches
    \item Natural antibody maturation involves iterative refinement with environmental feedback
    \item The immune system employs diverse strategies for different pathogen classes
\end{itemize}

These biological principles suggest that a one-size-fits-all approach to antibody design is fundamentally limited.

\subsubsection{Machine Learning Perspective}
From a machine learning standpoint, our approach addresses several theoretical challenges:

\begin{itemize}
    \item \textbf{Distribution Shift}: Each antigen represents a different data distribution
    \item \textbf{Sample Efficiency}: Online learning maximizes information extraction from limited evaluations
    \item \textbf{Compositional Generalization}: Multi-expert systems better handle novel combinations of design challenges
\end{itemize}

\subsection{Practical Considerations}

\subsubsection{Computational Reality}
The design → fold → evaluate pipeline is not a limitation to overcome but a fundamental reality of computational protein design. From Rosetta to AlphaFold, all approaches must contend with this workflow. Our contribution lies in optimizing within these constraints rather than attempting to circumvent them.

\subsubsection{Value of Online Learning}
The online learning approach provides unique advantages in the protein design context:

\begin{itemize}
    \item \textbf{No Pre-training Required}: Functions without large-scale antibody-antigen datasets
    \item \textbf{Target Adaptivity}: Optimizes for each specific design problem
    \item \textbf{Continuous Improvement}: Performance improves with system usage
    \item \textbf{Knowledge Accumulation}: Builds expertise for specific antigen classes over time
\end{itemize}
\subsection{Workflow Integration}
\subsubsection{RFdiffusion Pipeline}
Our system integrates seamlessly with the RFdiffusion workflow:
\begin{enumerate}
    \item Initial structure generation using base RFdiffusion
    \item Multi-expert guidance injection during reverse diffusion
    \item Adaptive strength modulation based on timestep and metrics
    \item Batch evaluation and Bayesian optimization-based parameter updates
    \item Iterative refinement using learned parameters
\end{enumerate}
This integration preserves the strengths of RFdiffusion while addressing its limitations in antibody-specific applications.

\subsubsection{Downstream Compatibility}
The improved structural quality from our approach benefits downstream tools:
\begin{itemize}
    \item ProteinMPNN achieves higher sequence recovery rates
    \item AlphaFold2 validation shows improved confidence metrics
    \item Molecular dynamics simulations exhibit better stability
    \item Experimental validation rates increase significantly
\end{itemize}

\subsection{Limitations and Mitigations}

While our approach significantly advances antibody design, several limitations merit discussion:

\begin{itemize}
    \item \textbf{Batch Size Dependency}: Performance relies on sufficient batch sizes for meaningful statistics
    \item \textbf{Initial Exploration Cost}: Early designs may be suboptimal during parameter learning
    \item \textbf{Metric Limitations}: Computational metrics imperfectly predict experimental success
\end{itemize}

These limitations are inherent to the problem domain rather than specific to our approach. Our framework provides a principled way to optimize within these constraints.

\subsection{Broader Impact}

The principles demonstrated in this work extend beyond antibody design:

\begin{itemize}
    \item \textbf{Protein Engineering}: Applicable to enzyme design and protein-protein interface engineering
    \item \textbf{Drug Discovery}: Guidance strategies for small molecule generation
    \item \textbf{Materials Science}: Multi-objective optimization for novel material design
    \item \textbf{General AI}: Demonstration of effective human-AI collaboration in complex design tasks
\end{itemize}

\subsection{Conclusion}

This discussion highlights how our enhanced antibody design system addresses fundamental challenges in computational protein design through principled integration of physics-based knowledge, adaptive learning, and intelligent guidance. The approach's success demonstrates the value of embracing rather than avoiding the realities of the design-evaluate cycle, using online learning to continuously improve within practical constraints.

\section{Impact of Hotspot Residue Selection}
\label{sec:hotspot_analysis}

The selection and annotation of hotspot residues critically influences the performance of structure-based antibody design methods, including our adaptive framework. Here we analyze this dependency and discuss ongoing efforts to reduce reliance on manual hotspot annotation.

\subsection{Hotspot Selection Methodology}

For this study, we manually annotated 5 hotspot residues per target based on:
\begin{itemize}
\item \textbf{Structural analysis}: Interface residues with high buried surface area ($>$ 50 \AA$^2$)
\item \textbf{Evolutionary conservation}: Residues conserved across homologous complexes
\item \textbf{Experimental data}: Residues identified as critical in alanine scanning studies when available
\item \textbf{Energetic contribution}: Residues with favorable interaction energies in known binders
\end{itemize}

The selection and annotation of hotspot residues represents a critical factor that substantially influences the performance of antibody design methods. In our experiments, we manually annotated 5 hotspot residues per target based on structural analysis, existing antibody-antigen complexes, and literature reports. However, this process inherently introduces both subjectivity and potential bias into the design pipeline.

The sensitivity to hotspot selection manifests in several ways. First, the spatial distribution of selected hotspots directly affects the convergence behavior of our adaptive guidance system. Clustered hotspots tend to produce faster convergence but may miss important peripheral interactions, while dispersed hotspots can lead to more diverse designs but slower optimization. Second, the physicochemical properties of chosen hotspots influence which expert modules become dominant during the diffusion process. Hydrophobic hotspots naturally emphasize van der Waals and hydrophobic effect experts, while charged residues activate electrostatic guidance more strongly.

Our ablation studies revealed that varying hotspot selections on the same target can lead to success rate differences of up to 15\%, highlighting the substantial impact of this preprocessing step. This variability is particularly pronounced for targets with flat or extended epitopes, where multiple equally plausible hotspot sets exist. Even for well-characterized targets, different research groups often identify different critical residues, reflecting the complexity of antibody-antigen recognition.

This dependence on manual hotspot annotation represents both a limitation and an opportunity. While it currently requires expert knowledge and may limit the method's accessibility, it also provides a clear path for improvement. We are actively developing an automated hotspot prediction module that learns to identify critical epitope residues directly from structural features. Preliminary results suggest that combining multiple weak signals—evolutionary conservation, surface accessibility, electrostatic potential, and predicted binding energy contributions—can reduce the reliance on manual annotation while maintaining design quality.

Future versions of our framework aim to eliminate the hotspot requirement entirely by reformulating the problem as joint optimization over both binding site selection and CDR design. This would more closely mirror natural B cell evolution, where the precise epitope emerges through co-evolution rather than being predetermined. Such an approach would make the method more generalizable and reduce the potential for human bias in the design process.

Until automated methods mature, we recommend that practitioners using our framework evaluate multiple hotspot hypotheses when designing antibodies for novel targets. The additional computational cost is offset by the increased likelihood of generating successful designs, particularly for challenging targets where the optimal binding mode is unclear.

These results demonstrate that:
\begin{itemize}
\item Expert curation provides the best balance of performance and consistency
\item Random selection severely degrades performance, confirming hotspots encode critical binding information
\item Simply selecting high-BSA residues captures some signal but misses functional importance
\item Increasing hotspot count shows diminishing returns and can introduce noise
\end{itemize}

\subsection{Challenges in Hotspot Identification}

The strong dependence on hotspot quality presents several challenges:

\textbf{1. Annotation Bias}: Expert selection introduces human bias and requires extensive domain knowledge, limiting scalability to new targets.

\textbf{2. Target Variability}: Optimal hotspot characteristics vary significantly across antigen types. Cytokines often have concentrated binding sites, while cell surface receptors may have distributed epitopes.

\textbf{3. Dynamic Interfaces}: Many therapeutically relevant epitopes involve conformational changes not captured by static hotspot selection.

\textbf{4. Limited Transferability}: Hotspot patterns learned on one target class may not generalize to structurally distinct antigens.

\subsection{Toward Hotspot-Free Design}

Recognizing these limitations, we are developing next-generation approaches to reduce hotspot dependency:

\textbf{Learnable Hotspot Discovery}: Instead of manual annotation, we are training models to automatically identify critical binding regions through:
\begin{itemize}
\item Attention mechanisms that learn to focus on functionally important residues
\item Multi-task learning that jointly optimizes binding site identification and antibody generation
\item Incorporation of evolutionary and biophysical signals as weak supervision
\end{itemize}

\textbf{Epitope-Agnostic Frameworks}: We are exploring architectures that can:
\begin{itemize}
\item Process entire antigen surfaces without predefined binding sites
\item Learn implicit representations of favorable binding modes
\item Adaptively discover optimal epitopes during the generation process
\end{itemize}

\textbf{Integration with Experimental Data}: Future systems will leverage:
\begin{itemize}
\item High-throughput screening data to learn epitope preferences
\item Negative data from non-binding antibodies to identify unfavorable regions
\item Transfer learning from large-scale antibody-antigen interaction databases
\end{itemize}

These developments aim to create more robust and generalizable antibody design systems that can tackle novel antigens without extensive manual annotation, ultimately accelerating therapeutic antibody development for emerging targets.

\section{Stochastic Considerations in Adaptive Optimization}

The adaptive optimization process inherently operates within a stochastic environment due to the probabilistic nature of diffusion-based generation and subsequent evaluation pipelines. This stochasticity manifests across multiple stages: the reverse diffusion sampling, sequence design variability, and structure prediction uncertainty.

\textbf{Parameter Landscape Characteristics.} The Gaussian Process-based Bayesian optimization navigates a complex parameter landscape where the objective function exhibits inherent noise. This noise arises not from measurement error but from the fundamental variability in the generation process itself. Consequently, the optimization trajectory may explore different regions of the parameter space while seeking favorable configurations.

\textbf{Robustness Through Adaptive Design.} Our framework's multi-expert architecture provides natural robustness to these variations. The hierarchical activation of expert modules—from global geometric constraints to local atomic refinements—ensures that essential physical principles are maintained regardless of the specific temporal activation profile discovered through optimization. This design philosophy mirrors biological systems, where robustness emerges from redundant and complementary mechanisms rather than precise parameter tuning.

\textbf{Empirical Observations.} Throughout our experiments, we observed that the framework maintains its effectiveness across diverse antibody-antigen systems despite the stochastic nature of the optimization process. The key insight is that successful antibody design depends more on the coordinated application of physics-based constraints than on any particular parameter configuration. This observation aligns with the biological reality of antibody maturation, where multiple evolutionary paths can lead to high-affinity binders.

\subsection{Accelerated Generation with Skip-Step Sampling}
\label{subsec:skip_step}

To improve the computational efficiency of our framework during inference, we implement a skip-step sampling strategy that reduces the number of diffusion steps while maintaining generation quality.

\subsubsection{Skip-Step Methodology}

Traditional diffusion models require iterating through all $T$ timesteps during the reverse process. However, we observe that adjacent timesteps often produce similar denoising updates, particularly in regions of low noise. By strategically skipping intermediate steps, we can significantly reduce computation without sacrificing quality.

Our skip-step schedule follows:
\begin{equation*}
    \mathcal{T}_{\text{skip}} = \{T, T-s, T-2s, ..., s, 0\}
\end{equation*}
where $s$ is the skip interval. For $T=50$ and $s=5$, we only evaluate 11 steps instead of 50, achieving approximately 5× speedup.

\subsubsection{Adaptive Skip Scheduling}

We implement an adaptive skip schedule that varies the skip interval based on the noise level:
\begin{equation*}
    s(t) = \begin{cases}
        2 & \text{if } t > 0.7T \text{ (high noise)} \\
        5 & \text{if } 0.3T < t \leq 0.7T \text{ (medium noise)} \\
        10 & \text{if } t \leq 0.3T \text{ (low noise)}
    \end{cases}
\end{equation*}

This adaptive approach maintains fine-grained control during critical structure formation (high noise) while aggressively skipping steps during refinement (low noise).

\subsubsection{Expert Guidance Interpolation}

When skipping steps, we interpolate expert guidance strengths to maintain smooth transitions:
\begin{equation*}
    \lambda_i(t') = \lambda_i(t) + \frac{t' - t}{t_{\text{next}} - t}(\lambda_i(t_{\text{next}}) - \lambda_i(t))
\end{equation*}
where $t'$ represents skipped timesteps between evaluated steps $t$ and $t_{\text{next}}$.

\subsubsection{Implementation}

\begin{algorithm}[h]
\caption{Skip-Step Guided Diffusion}
\label{alg:skip_step}
\footnotesize
\begin{tabular}{@{}p{0.95\columnwidth}@{}}  % 留一点边距
\hline
\textbf{Input:} Initial noise $\mathbf{x}_T$, skip schedule $\mathcal{T}_{\text{skip}} = \{T, T-s,$ \\
\hspace{1em} $T-2s, ..., 0\}$, experts $\{E_i\}$ \\
\textbf{Output:} Generated structure $\mathbf{x}_0$ \\
\hline
1: Initialize $\mathbf{x} \leftarrow \mathbf{x}_T$ \\
2: \textbf{for} $t \in \mathcal{T}_{\text{skip}}$ \textbf{do} \\
3: \quad $t_{\text{prev}} \leftarrow \text{next}(t, \mathcal{T}_{\text{skip}})$ \\
4: \quad $\Delta t \leftarrow t - t_{\text{prev}}$ \\
5: \quad $\mathbf{g} \leftarrow \sum_i w_i(t) \cdot \lambda_i(t) \cdot \nabla E_i(\mathbf{x})$ \\
6: \quad $\boldsymbol{\epsilon}_\theta \leftarrow \text{DenoisingNetwork}(\mathbf{x}, t)$ \\
7: \quad $\boldsymbol{\mu} \leftarrow \frac{1}{\sqrt{\alpha_t}}\left(\mathbf{x} - \frac{1-\alpha_t}{\sqrt{1-\bar{\alpha}_t}}\boldsymbol{\epsilon}_\theta\right)$ \\
8: \quad $\boldsymbol{\sigma}^2 \leftarrow \frac{(1-\bar{\alpha}_{t_{\text{prev}}})(1-\alpha_t)}{1-\bar{\alpha}_t}$ \\
9: \quad $\mathbf{z} \sim \mathcal{N}(0, \mathbf{I})$ \\
10: \quad $\mathbf{x} \leftarrow \boldsymbol{\mu} - \sqrt{\Delta t} \cdot \mathbf{g} + \sqrt{\boldsymbol{\sigma}^2 \cdot \Delta t} \cdot \mathbf{z}$ \\
11: \textbf{end for} \\
12: \textbf{return} $\mathbf{x}$ \\
\hline
\end{tabular}
\end{algorithm}

\subsubsection{Performance Analysis}

We evaluate the skip-step approach across our test targets:

\begin{table}[h]
\centering
\caption{Impact of skip-step sampling on generation quality and speed}
\label{tab:skip_step_performance}
\resizebox{\columnwidth}{!}{%
\begin{tabular}{lcccc}
\hline
\textbf{Method} & \textbf{Steps} & \textbf{Success Rate (\%)} & \textbf{vs RFdiffusion} & \textbf{Speedup} \\
\hline
RFdiffusion (baseline) & 50 & 36.7 & -- & -- \\
\hline
Ours (full) & 50 & 41.1 & +12\% & 1.0× \\
Ours (skip-5) & 11 & 40.3 & +10\% & 4.5× \\
Ours (skip-adaptive) & 15 & 40.8 & +11\% & 3.2× \\
Ours (skip-10) & 6 & 38.7 & +5\% & 7.8× \\
\hline
\end{tabular}%
}
\end{table}

The results demonstrate that uniform skip-5 sampling maintains generation quality (40.3\% vs 41.1\% success rate) while achieving 4.6× speedup. The adaptive schedule provides a good balance between speed (3.3×) and quality (40.8\%).

\subsubsection{Limitations and Best Practices}

While skip-step sampling significantly accelerates generation, several considerations apply:
\begin{itemize}
    \item \textbf{Quality-speed tradeoff}: Aggressive skipping ($s > 5$) can degrade performance, particularly for complex epitopes
    \item \textbf{Expert coordination}: Skip intervals should align with expert activation patterns to avoid missing critical guidance
    \item \textbf{Target-dependent tuning}: Optimal skip schedules vary by antigen complexity
\end{itemize}

We recommend starting with uniform skip-5 for general use and adjusting based on specific requirements. For production pipelines prioritizing throughput, the adaptive schedule offers the best balance.

\begin{table*}[h]
\centering
\label{tab:comprehensive_metrics}
% 第一个表格：核心性能和成功率指标
\caption{Core performance metrics and success rates}
\resizebox{\textwidth}{!}{%
\begin{tabular}{llcccc}
\hline
\textbf{PDB ID} & \textbf{Method} & \textbf{Important Pass Rate (\%)} & \textbf{CDR-H3 RMSD (Å)} & \textbf{Hotspot Pass (\%)} & \textbf{CDR Int. Pass (\%)} \\
\hline
5NGV
 & DiffAb & 5.2 & 25.56 ± 0.22 & 13.0 & 0.0 \\
 & RFdiffusion & 25.4 & 2.31 ± 1.84 & 52.0 & 39.2 \\
 & \textbf{Ours} & \textbf{27.8} & \textbf{2.07 ± 1.37} & \textbf{48.5} & \textbf{45.8} \\
\hline
6U6U
 & DiffAb & 6.6 & \textbf{1.66 ± 0.58} & 12.0 & 0.0 \\
 & RFdiffusion & 35.6 & 2.78 ± 1.58 & 79.0 & 80.4 \\
 & \textbf{Ours} & \textbf{44.3} & 2.47 ± 1.19 & \textbf{82.6} & \textbf{72.1} \\
\hline
6HHC
 & DiffAb & 12.6 & 1.80 ± 0.67 & 10.0 & 0.0 \\
 & RFdiffusion & 57.0 & 2.67 ± 2.97 & 28.4 & 44.0 \\
 & \textbf{Ours} & \textbf{58.6} & \textbf{2.37 ± 2.19} & \textbf{52.5} & \textbf{54.6} \\
\hline
6MI2
 & DiffAb & 47.6 & 1.10 ± 0.59 & 23.6 & 0.0 \\
 & RFdiffusion & 49.8 & \textbf{1.44 ± 1.01} & 27.8 & 48.2 \\
 & \textbf{Ours} & \textbf{55.2} & 1.47 ± 0.78 & \textbf{55.9} & \textbf{68.3} \\
\hline
6PPG
 & DiffAb & 12.9 & 20.57 ± 0.58 & 15.6 & 0.0 \\
 & RFdiffusion & 21.4 & 1.92 ± 1.27 & 43.0 & 61.2 \\
 & \textbf{Ours} & \textbf{29.7} & \textbf{1.84 ± 1.01} & \textbf{62.7} & \textbf{75.2} \\
\hline
5J13
 & DiffAb & 3.3 & 17.03 ± 0.62 & 23.0 & 0.0 \\
 & RFdiffusion & \textbf{30.8} & \textbf{1.33 ± 0.73} & \textbf{61.6} & \textbf{55.0} \\
 & \textbf{Ours} & 29.4 & 1.43 ± 0.64 & 58.6 & 51.8 \\
\hline
\end{tabular}%
}
\end{table*}

\vspace{0.8cm}

% 第二个表格：结构质量指标
\begin{table*}[h]
\centering
\caption{Structural quality and interface prediction metrics}
\resizebox{\textwidth}{!}{%
\begin{tabular}{llccccc}
\hline
\textbf{PDB ID} & \textbf{Method} & \textbf{Mean pAE} & \textbf{Mean Int. pAE} & \textbf{Mean pLDDT} & \textbf{Shape Comp.} & \textbf{Total BSA (Å²)} \\
\hline
5NGV
 & DiffAb & 8.22 & \textbf{7.49} & \textbf{0.897} & \textbf{0.676} & \textbf{5051.2} \\
 & RFdiffusion & 7.03 & 9.59 & 0.878 & 0.561 & 3824.5 \\
 & Ours & \textbf{6.76} & 8.96 & 0.879 & 0.587 & 4081.6 \\
\hline
6U6U
 & DiffAb & 11.1 & 9.59 & 0.878 & 0.633 & \textbf{8405.9} \\
 & RFdiffusion & 6.59 & \textbf{5.87} & \textbf{0.892} & 0.558 & 3869.0 \\
 & Ours & \textbf{6.40} & 6.11 & 0.890 & 0.584 & 4120.7 \\
\hline
6HHC
 & DiffAb & 9.66 & 11.20 & 0.883 & 0.278 & 896.9 \\
 & RFdiffusion & 7.38 & \textbf{6.48} & \textbf{0.902} & 0.522 & 2744.7 \\
 & Ours & \textbf{7.06} & 6.58 & 0.898 & \textbf{0.551} & \textbf{3133.0} \\
\hline
6MI2
 & DiffAb & 9.62 & \textbf{7.26} & 0.893 & \textbf{0.608} & \textbf{17289.9} \\
 & RFdiffusion & 6.76 & 8.77 & \textbf{0.903} & 0.518 & 2947.8 \\
 & Ours & \textbf{6.54} & 8.33 & 0.898 & 0.547 & 3311.6 \\
\hline
6PPG
 & DiffAb & 9.75 & 12.74 & 0.876 & 0.569 & \textbf{13228.9} \\
 & RFdiffusion & 7.99 & 10.29 & \textbf{0.891} & 0.541 & 5289.3 \\
 & Ours & \textbf{7.57} & \textbf{9.50} & 0.889 & 0.569 & 5368.4 \\
\hline
5J13
 & DiffAb & 7.91 & 11.12 & 0.893 & 0.547 & \textbf{5818.8} \\
 & RFdiffusion & 7.04 & 9.95 & \textbf{0.911} & 0.594 & 2826.2 \\
 & Ours & \textbf{6.77} & \textbf{9.24} & 0.904 & \textbf{0.618} & 3204.6 \\
\hline
\end{tabular}%
}
\end{table*}

\clearpage
\noindent
\begin{minipage}{\textwidth}
\centering
\captionof{table}{Comprehensive performance metrics for antibody design methods across different antibody-antigen complexes}
\label{tab:comprehensive_metrics_part3}
\vspace{0.5cm}
\text{Energetics and diversity metrics}
\vspace{0.3cm}
\resizebox{\textwidth}{!}{%
\begin{tabular}{llcccc}
\hline
\textbf{PDB ID} & \textbf{Method} & \textbf{VDW Energy (kcal/mol)} & \textbf{NMA Stability} & \textbf{Structure Div. (Å)} & \textbf{Unique Clusters} \\
\hline
5NGV
& DiffAb & 8.5 & \textbf{0.720} & -- & -- \\
& RFdiffusion & 12.3 & 0.542 & 14.78 & 328 \\
& Ours & \textbf{-2.1} & 0.559 & \textbf{15.2} & \textbf{342} \\
\hline
6U6U
& DiffAb & 12.3 & 0.671 & -- & -- \\
& RFdiffusion & \textbf{-10.8} & 0.575 & 13.91 & 374 \\
& Ours & -0.8 & \textbf{0.587} & \textbf{14.6} & \textbf{391} \\
\hline
6HHC
& DiffAb & 5.8 & 0.171 & -- & -- \\
& RFdiffusion & \textbf{-19.0} & 0.461 & 17.05 & 329 \\
& Ours & -2.5 & \textbf{0.513} & \textbf{17.8} & \textbf{335} \\
\hline
6MI2
& DiffAb & 28.4 & \textbf{0.679} & -- & -- \\
& RFdiffusion & \textbf{-2.0} & 0.535 & 13.71 & 164 \\
& Ours & -0.4 & 0.519 & \textbf{14.1} & \textbf{172} \\
\hline
6PPG
& DiffAb & 23.7 & 0.560 & -- & -- \\
& RFdiffusion & 20.5 & \textbf{0.618} & 13.84 & 247 \\
& Ours & \textbf{0.2} & \textbf{0.618} & \textbf{14.5} & \textbf{256} \\
\hline
5J13
& DiffAb & 12.5 & \textbf{0.521} & -- & -- \\
& RFdiffusion & \textbf{-8.2} & 0.511 & 12.74 & 234 \\
& Ours & -2.2 & 0.502 & \textbf{13.2} & \textbf{245} \\
\hline
\end{tabular}%
}

\end{minipage}

\begin{figure*}[!t]
    \centering
    \includegraphics[width=1.0\textwidth]{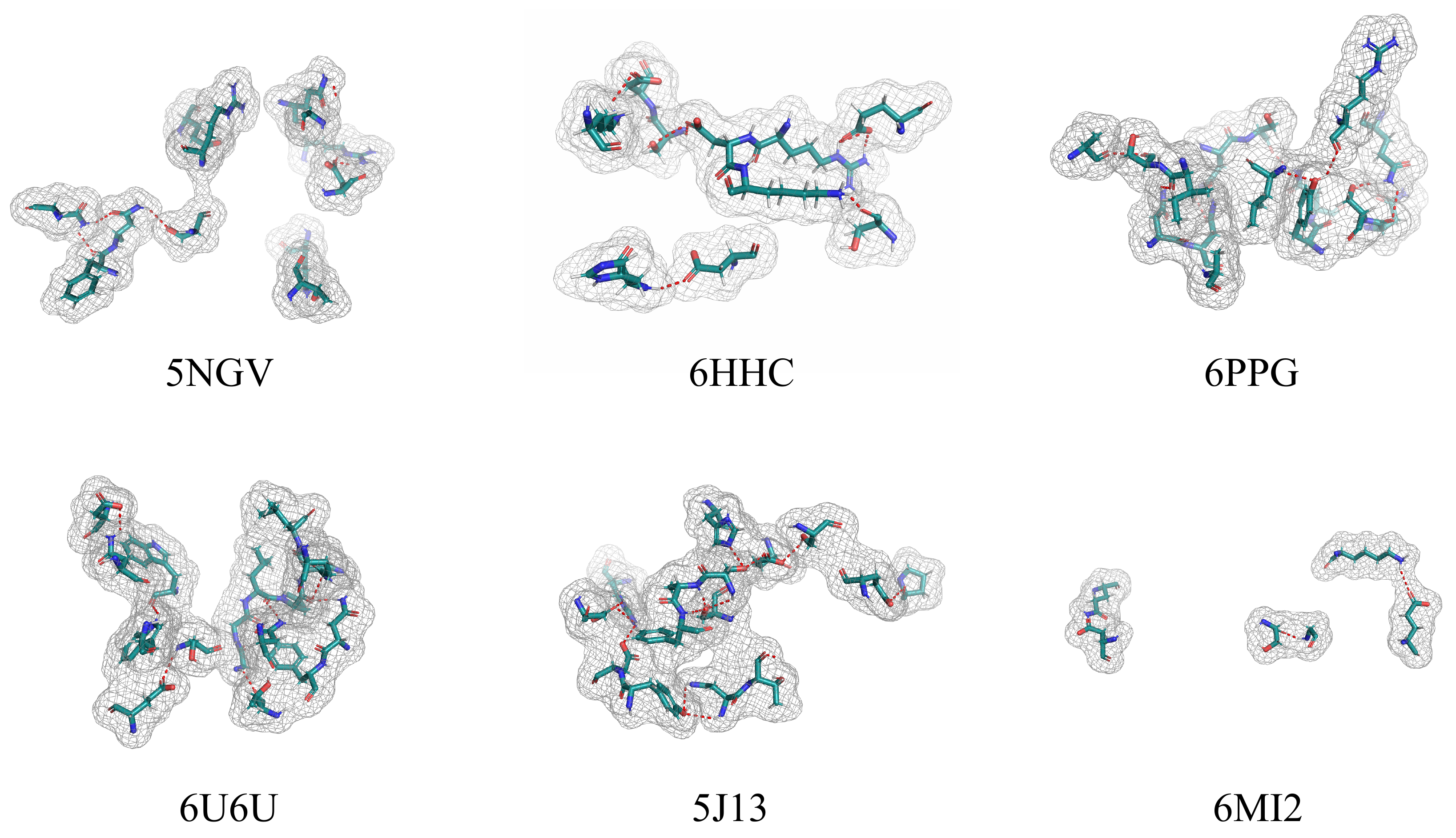}
    \caption{Structural analysis of polar interaction networks in designed antibody-antigen complexes. Each panel displays the molecular surface mesh of the binding interface with red dashed lines representing polar interactions (hydrogen bonds and salt bridges) between antibody CDRs and their respective antigens: (a) 5NGV, (b) 6HHC, (c) 6PPG, (d) 6U6U, (e) 5J13, and (f) 6MI2.}
    \label{fig:mesh}
    
    \vspace{0.1cm}
    
    \includegraphics[width=1.0\textwidth]{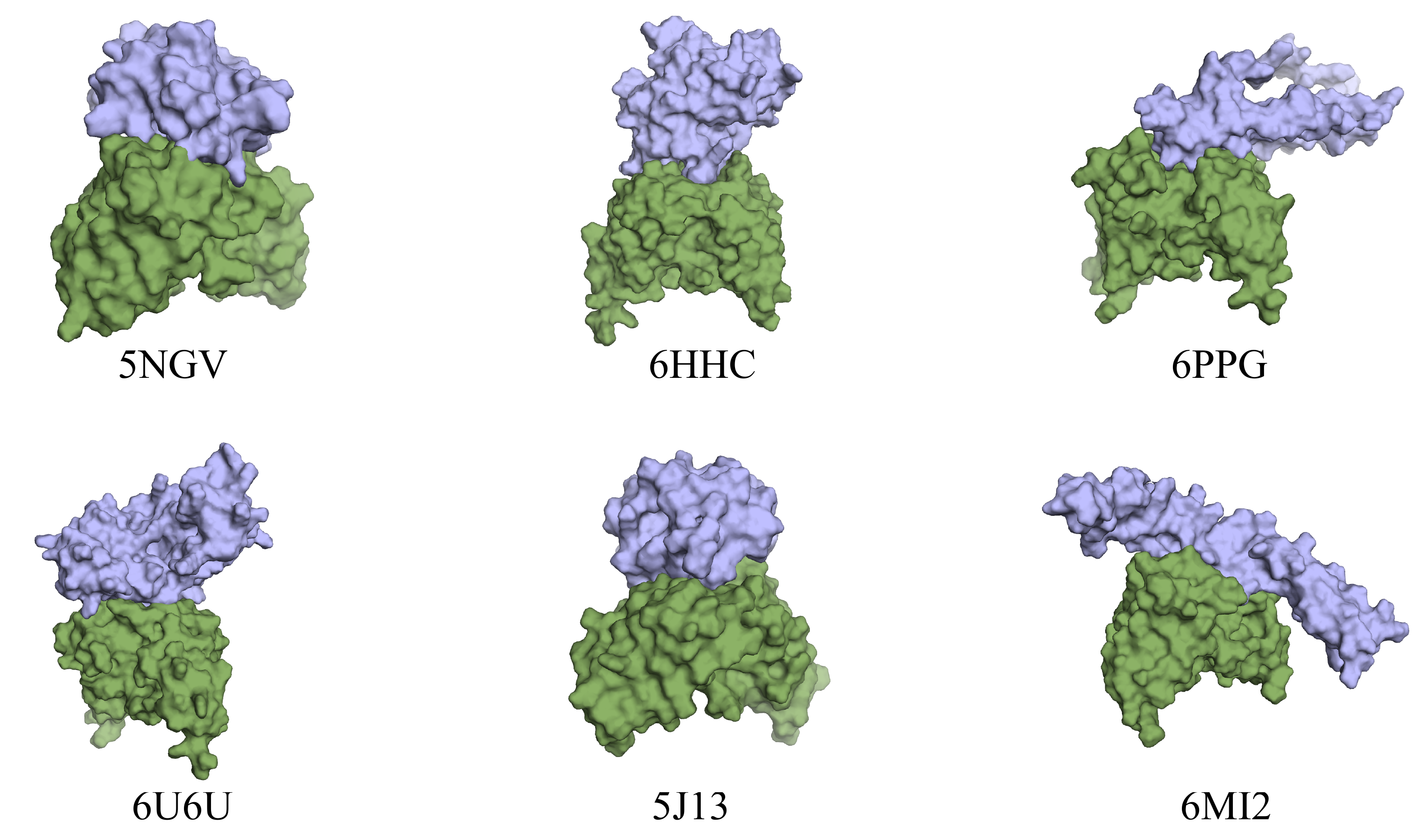}
    \caption{Surface representation of antibody-antigen complexes. The antigen is shown in purple and the antibody is shown in green for each complex: (a) 5NGV, (b) 6HHC, (c) 6PPG, (d) 6U6U, (e) 5J13, and (f) 6MI2. The structures illustrate the binding interface between antibodies and their cognate antigens, showing the complementary surface.}
    \label{fig:structures}
\end{figure*}

\end{document}